\documentclass[10pt,journal,compsoc]{IEEEtran}
\usepackage{graphicx}
\usepackage{float}
\usepackage{caption}
\usepackage{comment}
\usepackage{amssymb}
\usepackage{amsmath,amssymb} 
\usepackage{color}
\usepackage{multirow}
\usepackage{booktabs}
\usepackage{xspace}
\usepackage{makecell}
\usepackage{cite}
\usepackage{url}
\usepackage{subcaption}
\usepackage[mathscr]{eucal}

\ifCLASSINFOpdf
\else
\fi
\begin{document}
%
\title{CycleSegNet: Object Co-Segmentation with Cycle Refinement and Region Correspondence}
%
%
%
%

\author{Chi Zhang$^*$\thanks{$^*$ indicates equal contribution.},
        Guankai Li$^*$,
        Guosheng Lin\thanks{Corresponding author: Guosheng Lin.},
        Qingyao Wu,
        Rui Yao
\IEEEcompsocitemizethanks{\IEEEcompsocthanksitem  Chi Zhang, Guankai Li, Guosheng Lin are with School of Computer Science and Engineering, Nanyang Technological University (NTU), Singapore 639798 (email: guankai001@e.ntu.edu.sg, chi007@e.ntu.edu.sg, gslin@e.ntu.edu.sg).

\IEEEcompsocthanksitem
Qingyao Wu is with School of Software Engineering, South China University of Technology, and Pazhou Lab, Guangzhou, China (email: qyw@scut.edu.cn)

\IEEEcompsocthanksitem
R. Yao is with the School of Computer Science and Technology, China University of Mining and Technology, Xuzhou 221116, China (e-mail: ruiyao@cumt.edu.cn)

}

}

\IEEEtitleabstractindextext{%

\begin{abstract}

Image co-segmentation is an active computer vision task that aims to segment the common objects from a set of images. Recently, researchers design various learning-based algorithms to undertake the co-segmentation task. The main difficulty in this task is how to effectively transfer information between images to make conditional predictions. In this paper, we present CycleSegNet, a novel framework for the co-segmentation task. Our network design has two key components: a region correspondence module which is the basic operation for exchanging information between local image regions, and a cycle refinement module, which utilizes ConvLSTMs to progressively update image representations and exchange information in a cycle and iterative manner. Extensive experiments demonstrate that our proposed method significantly outperforms the state-of-the-art methods on four popular benchmark datasets --- PASCAL VOC dataset, MSRC dataset, Internet dataset, and iCoseg dataset, by 2.6\%, 7.7\%, 2.2\%, and 2.9\%, respectively.

\end{abstract}

}

\maketitle

\IEEEdisplaynontitleabstractindextext

%
\IEEEpeerreviewmaketitle

\section{Introduction}

Image co-segmentation is an active computer vision topic with a long research history, which aims to segment the common objects jointly from a set of images.
Image co-segmentation algorithms have shown their usages in various computer vision tasks, such as image retrieval~\cite{vicente2011object}, 3D reconstruction~\cite{mustafa2017semantically}, photo collections~\cite{rother2006cosegmentation}, image matching~\cite{chen2015co,zhang2020deepemd,zhang2020deepemdv2}, and video object tracking~\cite{rother2006cosegmentation,liu2020crnet,liu2020weakly}. 
Recently, data-driven deep neural networks based methods attract wide interest in the literature. 
The powerful deep neural networks along with the challenging evaluation benchmarks built upon large-scale public datasets have brought this task to a new era and make it more challenging.
Although deep neural networks have shown remarkable success in many other segmentation tasks, such as semantic segmentation~\cite{long2015fully,fu2019stacked,jing2019coarse,zhang2019pyramid,zhang2019canet,liu2020weakly,chen2020compositional}, interactive segmentation~\cite{sakinis2019interactive}, and instance segmentation~\cite{he2017mask,zhang2019mask,yu2019exemplar}, these models are not directly applicable to the co-segmentation tasks, as the outputs are conditioned on the pairwise or group-wise relations between input images.

\begin{figure}[t]
    \centering
      \includegraphics[width=0.90\linewidth]{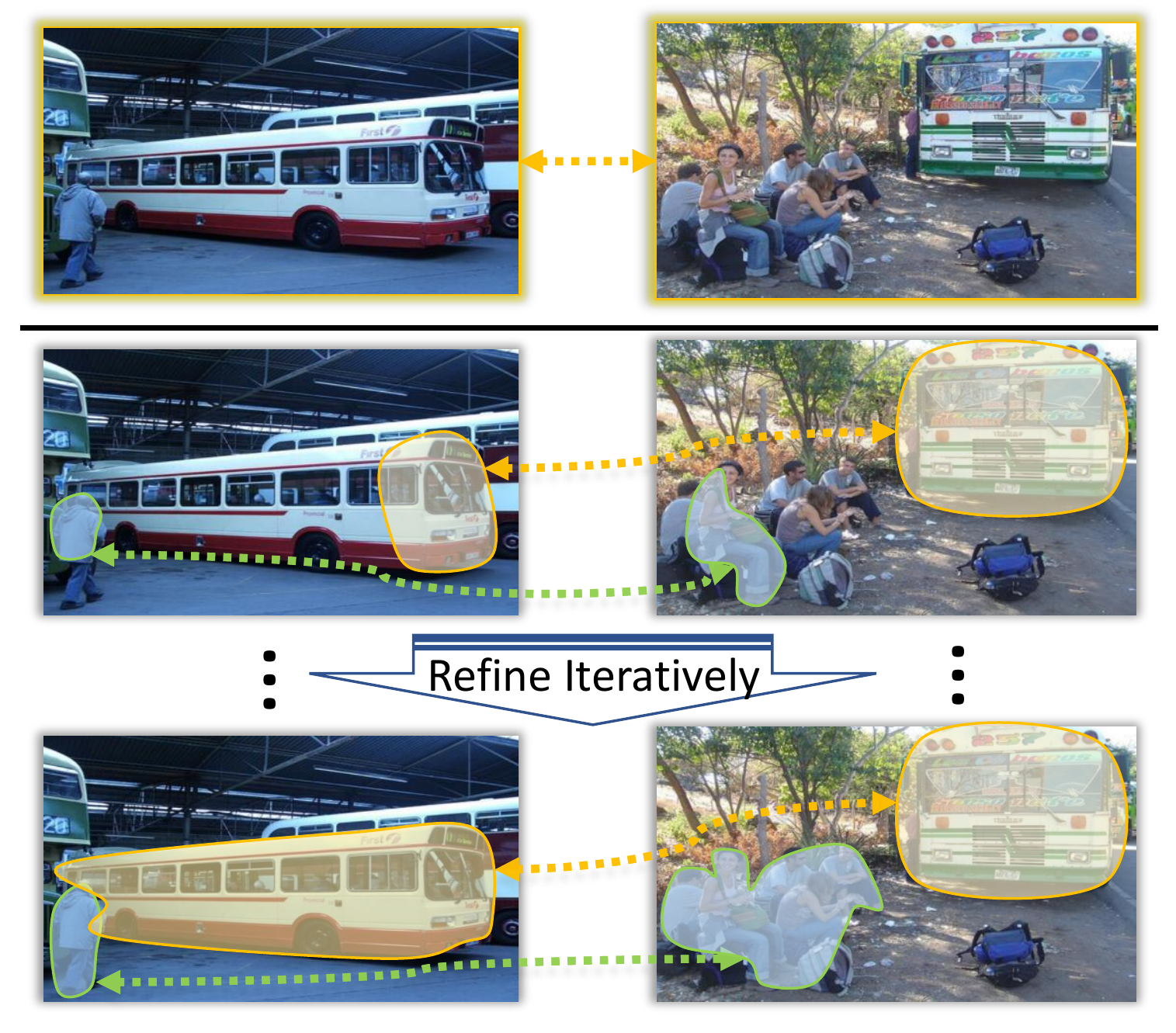}
    \caption{An illustration of our motivation. Previous methods exchange image-level semantics as cues for image co-segmentation (up), while our method establishes  correspondence between local image regions and refines the predictions iteratively (down).}
      \label{fig:teaser}
\end{figure}

Many existing CNN based methods~\cite{li2018deep,chen2018semantic,banerjee2019cosegnet,liu2020crnet}  solve the image co-segmentation problems by employing a pair of parameter-shared Siamese networks to generate feature representations of two images and using various methods to transfer information between them to make conditional predictions.
The intuitions behind these methods are that in order to segment the common objects between images, the cues about what objects are existing in individual images must be exchanged for predictions.
The challenge here is how to effectively transfer useful information based on image representations. 
Directly copying the image representations is unrealistic, as images have structured representations and no correspondence information is provided.
Hence, simply multiplying or concatenating structured image representations would not work when no alignment information is provided.
To tackle this problem, many recent works~\cite{banerjee2019cosegnet,chen2018semantic}  bypass the correspondence problem by squeezing the image structures and exchange information in the form of a  global image-level representation which is usually achieved by the global average pooling.
As a result, the prediction of each pixel location can refer to the same global representations that are exchanged.
However, squeezing the image structures into a global representation inevitably loses local discriminative information which can be useful to locate common objects from the scene. Moreover, as the image content often has a complex composition, including the cluttered background and objects from different classes, the image-level representation may also introduce noise into the cues for conditional predictions. 
In this paper, we propose to solve the aforementioned problems from two perspectives, which are illustrated in Figure~\ref{fig:teaser}.

First, we argue that a desired deep image co-segmentation algorithm should maintain the image structures and the exchanged information should contain local discriminative information as the cues for conditional predictions.
To achieve this goal, we design a region correspondence module (RCM) that utilizes the attention mechanism to establish the correspondence between local regional representations in different images. In the RCM, each pixel location can attend to its most relevant regions in other images, and thus the prediction of each pixel refers more to the relevant regions based on the attention values. 
In comparison with the cues generated by global average poolings, the cues in our method include more local discriminative features and contain less noise for prediction.

Second, we argue that the task of image co-segmentation can be modeled as an iterative optimization process where the predictions as well as the exchanged cues can be refined progressively. 
We take inspiration from how we humans locate common objects from two images --- given two images that contain the same objects, humans may observe two images alternatively and every time an image is watched, humans remember what exists in one image and try to find them in other images. Gradually, those exclusive image contents are out of consideration and the cues about the common objects are getting clearer.
Based on such intuition, we design a cycle refinement module (CRM) to refine feature representations for predictions in a cycle and iterative manner.
We employ a pair of parameter-shared convolutional LSTMs  (ConvLSTMs)~\cite{shi2015convolutional} to achieve this goal, which is operated at the bottleneck of an encoder-decoder network.
At each time step, the LSTM cell takes the exchanged information from the opposite branch as the input and updates its own embedding, which is then sent to the opposite branch as the exchanged cues in the next step. 
In this way, the cycle refinement module can continually exchange information between two images and refine the image representations. 
As a result, both the image representations for predictions and cues about the common objects can be refined iteratively, and they can benefit from each other in this process.
Specifically, a clear cue received from other images about the common objects can definitely improve the image representation that is used for conditional prediction, and at the same time, a good image representation that contains more information about the common objects and less about cluttered background produces better cues. 

An advantage of our design is that our network for co-segmentation can handle the input with paired images as well as a group of images with the same model and network parameters.
This is because the key operation in our network design is based on the attention mechanism and does not require a fixed number of pixel locations on each side. Therefore, we can transform the network-like relations between grouped inputs to a one-versus-rest relation to undertake the group segmentation task with the same model.

Finally, we develop a model variant that exploits multi-level features to better infer the common objects in two images. As is often observed in the CNN visualization literature~\cite{zeiler2014visualizing,liu2020weakly}, high-level features contain more class-related information while middle-level features correspond to object parts that may be shared across different categories. It is therefore useful to establish region connections at multiple semantic levels to improve predictions. We apply our proposed modules to the features at different stages independently and then fuse their results during the decoding process.
We show that our network can better locate the pixels belonging to the common objects by utilizing multi-level features.

To validate the effectiveness of our design, we implement various experiments on multiple datasets to investigate each component in our network. Extensive experiments demonstrate that our design  significantly outperforms  existing models and the baselines. The main contributions of this work are summarized as follows:
\begin{itemize}
\itemsep 0.1cm

    \item We propose a region correspondence module as a basic operation to exchange information between images for the co-segmentation task. The proposed module maintains local discriminative representations and utilizes the attention mechanism to directly establish correspondence between image regions.
    \item We propose a cycle refinement module that employs ConvLSTMs to progressively refine network predictions as well as the exchanged information. 
    \item We demonstrate that our model can handle inputs of paired images and grouped images with the same network and parameters.
    \item We develop a model variant that exploits multi-level features to undertake the co-segmentation task. The multi-level version of our network can better locate common objects and improve the performance.
    
	\item Experiments on four popular benchmarks --- PASCAL VOC dataset, MSRC dataset, Internet dataset, and iCoseg dataset, demonstrate that our proposed network on both co-segmentation and group segmentation tasks significantly outperforms the baseline methods and achieves new state-of-the-art performance with remarkable advantages.

\end{itemize}

\section{Related Work}
\textbf{Object co-segmentation.}
Early works often address object co-segmentation problems by comparing the visual features of paired images, such as foreground color histogram~\cite{rother2006cosegmentation}, SIFT~\cite{mukherjee2011scale}, and saliency~\cite{chang2011co}. 
 Joulin \textit{et al.}~\cite{joulin2010discriminative} show that finding discriminative features can be well adapted to solve the object co-segmentation tasks. 
Rubio \textit{et al.} \cite{rubio2012unsupervised} train SVM classifiers based on a Gaussian Mixture Model to capture the correspondence between different regions of input images.
Rubinstein \textit{et al.}\cite{rubinstein2013unsupervised} utilize dense correspondences and visual saliency to find out the sparsity and visual variability of the common objects in the whole image database.

In the past few years, with the assistance of deep learning, some CNN-based networks have been proposed to solve the co-segmentation task.
For instance, Quan \textit{et al.}\cite{quan2016object} propose a manifold ranking algorithm by combining the features obtained from the VGG network and handcrafted features for superpixels to implement object co-segmentation. 
Li \textit{et al.}\cite{li2018deep}  utilize a pure fully convolutional neural network to solve the object co-segmentation task. 
Their method has a Siamese encoder-decoder architecture and utilizes a mutual correlation layer to segment the common objects in a pair of images. 
Chen \textit{et al.}\cite{chen2018semantic} propose a semantic attention layer to spotlight feature channels, which is the first work that leverages the channel attentions for object co-segmentation tasks.
Zhang \textit{et al.}\cite{zhang2019deep} design a spatial modulator for group-wise mask learning and a semantic modulator for co-category classification.
Li \textit{et al.}\cite{li2019group} propose a co-attention recurrent unit to generate the group representation containing the synergetic information, which is broadcasted to each individual image to facilitate the inferring of common objects. 
Hsu \textit{et al.}\cite{hsu2018co} explore an unsupervised co-segmentation setting  where no mask supervision is provided, which is different from our task setting where the ground-truth masks are provided. To enable the learning of the deep neural networks without mask labels, they design two loss functions that aim to minimize inter-image object discrepancy and maximize intra-image figure-ground separation.
Image degradation is a common problem when an image segmentation model is applied to real world situations, while models trained with clean images may generate 
poor predictions. To solve this problem, Guo~\textit{et al.}~\cite{guo2019degraded} propose a novel Dense-Gram Network based on the Gram matrix to effectively reduce the gap.

\textbf{Iterative refinement in segmentation.}
Many previous works use iterative designs to optimize segmentation results. For example, McIntosh \textit{et al.} \cite{mcintosh2018recurrent} adopt convolutional LSTMs to control computation budgets by adjusting the number of iteration steps.
Wang \textit{et al.} \cite{wang2018weakly} propose to use saliency maps to iteratively refine segmentation results under the weakly supervised setting. 
Lin \textit{et al.} ~\cite{lin2016scribblesup} propose to use graphical models to optimize the mask learned by scribble supervision for semantic segmentation.
In \cite{hu2019attention} and \cite{zhang2019canet}, iterative structures are also used to improve few-shot segmentation results.
Compared with previous iterative structures, the main difference in our design is that our two branches can benefit from each other and meanwhile refine their own predictions in a cycle manner.  

\textbf{Co-saliency detection.} Co-saliency detection~\cite{wei2017group,zhang2015cosaliency,fan2021re,han2017unified,zhang2016co,zhang2016detection} is a closely related topic which aims to identify the common and salient objects from a group of images.
Zhang \textit{et al.}~\cite{zhang2015cosaliency} explore intra-saliency prior transfer and deep inter-saliency mining for co-saliency detection. 
Wei \textit{et al.}~\cite{wei2017group} introduce the group-wise feature representation learning and the collaborative learning to address the co-salient object discovery problem. 
To enable fast common information learning, Fan \textit{et al.}~\cite{fan2021re} propose a network to simultaneously embed the appearance and semantic features through a co-attention projection strategy. 
Zhang~\textit{et al.}~\cite{zhang2019rgb} utilize multi-level CNN features to improve the RGB-T salient object detection, which shares similar spirits with our model variants.
Li~\textit{et al.}~\cite{li2020asif} employ attention mechanisms to  integrate cross-modal and cross-level complementarity from multi-modal data for RGB-D salient object detection. We also utilize attention mechanisms in our design, but the attention is used to establish regional connections between images, which has a different purpose.
Zhang~\textit{et al.}~\cite{zhang2020revisiting} design a feature fusion network to undertake the RGB-T saliency detection task, which includes multi-scale, multi-modality, and multi-level feature fusion modules.
Yu~\textit{et al.}~\cite{yu2018co} explore a novel problem setting that aims to identify
common saliency within an image and propose a bottom-up framework to address the problem.
Compared with co-saliency detection, object co-segmentation must identify objects from multiple categories from a complex scene, which are not necessarily the salient areas in images. For example, the buses and the people in Figure~\ref{fig:teaser} are both common objects in the images.

\section{Method}

\begin{figure*}[t]
    \centering
      \includegraphics[width=0.95\linewidth]{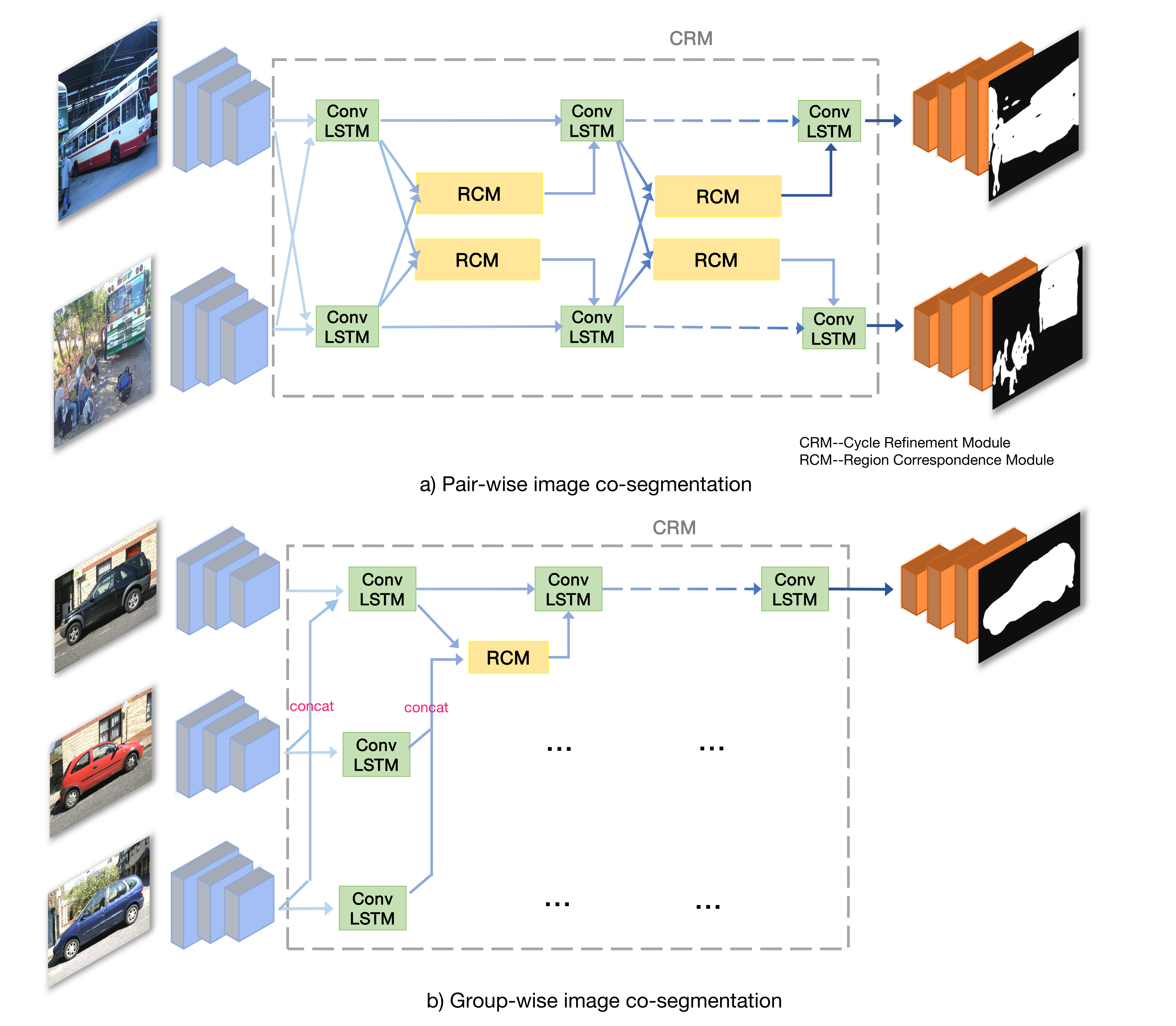}
    \caption{Our network for image co-segmentation with paired (a) or grouped (b) input images. The main components in our network include a Cycle Refinement Module (CRM) for feature updating  and a Region Correspondence Module (RCM) for information exchange.}
      \label{fig:crm-pair}
\end{figure*}

\begin{figure*}[t]
    \centering
      \includegraphics[width=0.80\linewidth]{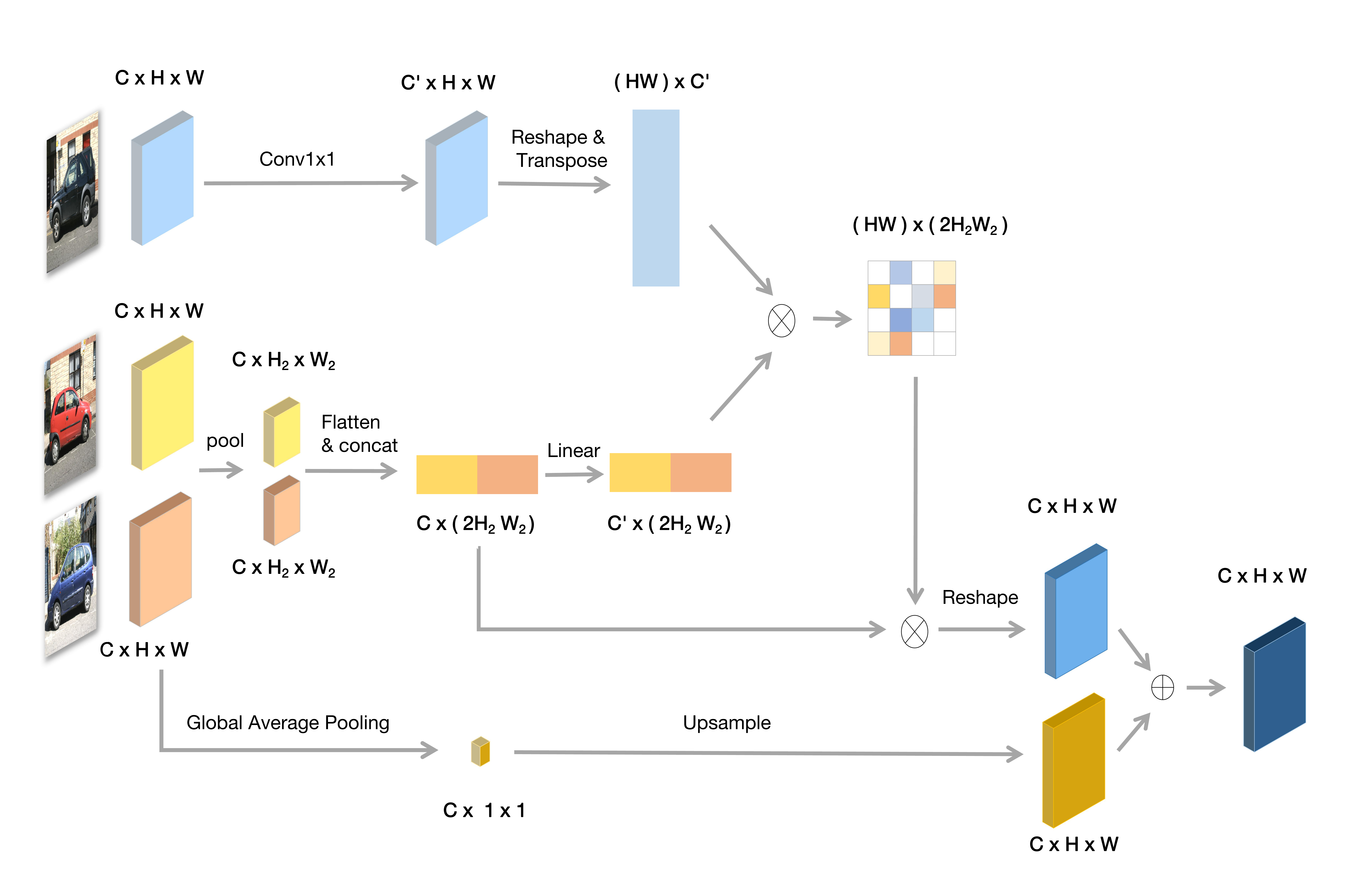}
    \caption{Architecture of the Region Correspondence Module. $\bigoplus$ indicates element-wise addition and $\bigotimes$ indicates matrix multiplication.}
      \label{fig:cross}
\end{figure*}

In this section, we present our framework for the image co-segmentation task. We begin with the description of our model in the case of paired input images.
Each image is encoded and decoded with a parameter-shared Siamese network. The network design has an encoder-decoder structure, and the two branches exchange information at the network bottlenecks. We first describe our model version that only exploits the features at the last layer of the encoder for information exchange, which is shown in Figure~\ref{fig:crm-pair}.  Then, we describe how to extend our network to the group-segmentation task without learning new parameters. Finally, we describe the model variant that uses multi-level features.

\subsection{Cycle Refinement Module}
The cycle refinement module (CRM) aims to update the image embeddings by incorporating information from the other image. The overall structure of the cycle refinement module can be found in Figure~\ref{fig:crm-pair}. In each step, the CRM  implicitly compares the co-occurrent semantics of two images and lets the representations focus on common objects progressively. To achieve this goal, we employ a ConvLSTM at the network bottleneck to undertake the tasks of information exchange and representation updating.  LSTM has proven its success in numerous computer vision and NLP tasks. Its feedback connections and memory cells make it well-suited for tasks with sequential inputs. The ConvLSTM further replaces the linear transformations in  LSTM with convolutional kernels, such that the LSTM owns large field-of-views and is suited for processing image data. A typical ConvLSTM cell has the following structures: 
\begin{align}
& i_{t}=\sigma (W_{xi}*X_{t}+W_{hi}*H_{t-1}+b_{i}),  \\
& f_{t}=\sigma (W_{xf}*X_{t}+W_{hf}*H_{t-1}+b_{f}), \\
& o_t=\sigma(W_{xo} * X_t+W_{ho} *H_{t-1}  +b_o), \\
& \widetilde{C_{t}}=i_{t}\odot tanh(W_{xc}*X_{t}+W_{hc}*H_{t-1}+b_{c}) \\
& C_{t}= f_{t}\odot C_{t-1}+i_{t}\odot\widetilde{C_{t}}  , \\
& H_{t}=o_{t} \odot tanh(C_{t}), 
\end{align}
where ${i_t}$ is the input gate, which controls the activation of the new input information; 
 ${f_t}$ is the forget gate that clears the past cell status; 
${o_t}$ is the output gate, controlling whether the latest cell output ${C_t}$ will be propagated to the final state ${H_t}$; 
$W$ is convolutional kernels;
$*$ denotes the convolution operator and $\mathcal{\odot}$ denotes the Hadamard product.
At each time step $t$, the ConvLSTM cell takes in an input $X_t$ and updates the hidden states $H_t$ and cell state $C_t$, which are both initialized by the image representations generated by the encoders.   The updated cell state $C_t$ is then sent to the region correspondence
module (described in Section~\ref{sec:crm}), which compares the cells from both sides and generates the inputs of the next step:
\begin{align}
 & X _{t}^{A}=\text{RCM} \left( C_{t-1}^{A},C_{t-1}^{B} \right), \\
 & X _{t}^{B}=\text{RCM} \left( C_{t-1}^{B},C_{t-1}^{A} \right).
\end{align}
At the last step, the hidden state $H$ is decoded by the decoder and generates the predicted mask of the common objects.
Based on this recurrent process, both the quality of the transferred representations and the accuracy of predictions can be improved gradually.
The number of steps for information exchange is flexible. In our experiment, we investigate the influence of the step numbers on the performance and the computation cost.
Our proposed CRM for image co-segmentation has a symmetric structure design and all the operations on both sides are parameter-shared.

\begin{figure*}[]
	\centering
	\includegraphics[width=0.95\linewidth]{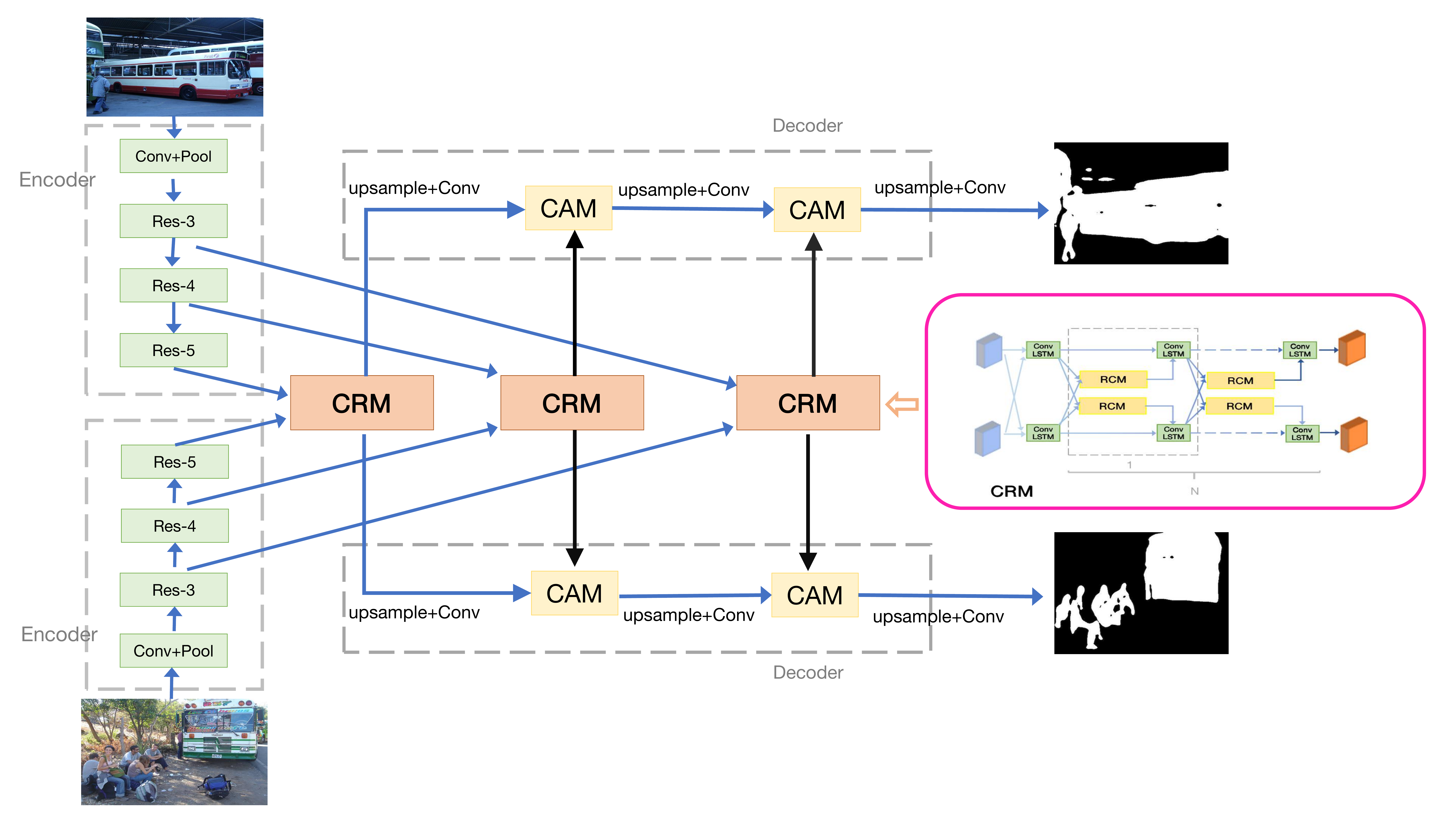}
	\caption{Our model variant utilizing multi-level features in the encoder for image co-segmentation. We apply the \textbf{CRM} module (described in Section~\ref{sec:crm} and illustrated in Figure~\ref{fig:crm-pair}) to features from different layers independently and fuse the resulting features using the Channel Attention module (CAM) proposed in~\cite{yu2018learning} during the upsampling stage.  }
	\label{fig:multi}
\vskip -1em
\end{figure*}

 \subsection{Region Correspondence Module}
 \label{sec:crm}
As we have seen in the cycle refinement module, the key component that exchanges information between two images is the region correspondence module.
Previous works often achieve this goal through a global representation as the exchanged information --- the exchanged global representation is then fused with the dense image representations by multiplication or concatenation. However, as image segmentation is a dense prediction task, local discriminative representations are important information to inference the common object regions. Therefore, we maintain the local representations of images and use attention mechanisms to establish the regional connections between two images. The architecture of RCM is illustrated in Figure~\ref{fig:cross}.  Specifically, the input to the region correspondence module is the current representations of two images $ C_{t}^{A}\in \mathbb{R}^{H_A\times W_A\times C}$ and  $ C_{t}^{B}\in \mathbb{R}^{H_B\times W_B\times C}$, which are the cell states in the ConvLSTMs.
When we want to transfer information from image $I_B$ to image $I_A$, we first get the regional representations of image $I_B$ by applying the ROI pooling to the representation $C_t^B$ which downsamples the image representations to a fixed spatial size $\hat{H} \times \hat{W}$:
\begin{equation}
\label{eq:roi}
    C_{t}^{'B}=\text{ROI}(C_{t}^{B}).
\end{equation}
The regional representations can provide context information of local regions, which are proven useful in many previous segmentation works~\cite{zhao2017pyramid}. We apply both the ROI average poling and the ROI max pooling to $C_{t}^{B}$, and fuse their results by convolutions to get $C_{t}^{'B}$.
The operations here share some similarities with the Channel Attention Block proposed in~\cite{woo2018cbam}, where the global average pooling and global max pooling are used together to generate channel attentions.
Then we compute the similarity between each feature point in  $C_{t}^{A}$ and $C_{t}^{'B}$ by dot product, which is implemented in parallel by matrix multiplication:
\begin{equation}
\label{eq:sim}
    S_t^{AB}=W^A(C_{t}^{A})  (W^B(C_{t}^{'B}))^T,
\end{equation}
where $S_t^{AB} \in \mathbb{R}^{H^AW^A \times \hat{H}\hat{W}}$, and W is the linear transformation function followed by ReLU non-linearity, implemented as $1\times1$ convolutions. For notational convenience, we omit the reshape operations in Equation~\ref{eq:sim}.
Intuitively, the function $W$ projects the feature representations into a space for computing feature similarity and the affinity matrix $ S_t^{AB}$ can reflect the correlations between local representations. Then, we normalize the affinity matrix by softmax and use the normalized affinity matrix to query features from the regional representations of image $I^B$:
\begin{equation}
    X^{AB}_t=\text{softmax}(S_t^{AB})  C_{t}^{'B}, 
\end{equation}
where $ X^{AB}_t\in \mathbb{R}^{H_AW_A\times C}$.
We also incorporate the global statistics of image $I^B$ into the exchanged information by global average pooling, and upsample it to the same spatial size of image $I^A$. The final output of  region correspondence module is
\begin{align}
 \text{RCM} \left( C_{t}^{A},C_{t}^{B} \right)= (X^{AB}_t+ \text{Up}(\text{AvgPool}(C_{t}^{B})))/2.
\end{align}
When transferring information from image $I^A$ to image $I^B$, we can simply swap the notation of $A$ and $B$ in the above equations, and apply the same operations.
As we can see above, the exchanged information for each pixel location is different. Every pixel can attend all the local regions in the other image and selectively extract information from different regions.

\subsection{Group Segmentation}
Group segmentation is a special case of the co-segmentation task, where the goal is to segment the common objects from a group of images. An advantage of our network is that the image co-segmentation model for paired input images can also be applied to the group segmentation task with only minor modifications and without learning new parameters. 
Since all the operations in the region correspondence module do not require a fixed spatial size of the representations, we can use the region correspondence module to directly transfer information from all other images to the target image by treating all other images as a big virtual image.
Figure~\ref{fig:crm-pair} (b) illustrates how our co-segmentation network can be used to undertake the group segmentation task. 
For example, when a group segmentation task has three input images, the prediction of an image should refer to information from the other two images. We achieve this goal by constructing an affinity matrix that contains the similarity between the regional representations in the target image and all other images. We can therefore selectively extract information from all other images based on the attention distribution. 
Specifically, the Equation~\ref{eq:roi} and Equation~\ref{eq:sim} become:
\begin{align}
    &C_{t}^{'BC}=\text{Flatten}(\text{ROI}(C_{t}^{B}))||\text{Flatten}(\text{ROI}(C_{t}^{C})),\\
    &S_t^{A\_BC}=W^A(C_{t}^{A}) (W^B(C_{t}^{'BC}))^T,
\end{align}
where Flatten is the operation that reshapes the 2D tensor to 1D tensor and  $||$ denotes the concatenation operator. To make it more clear, the shapes of the tensors above are listed here: $C_{t}^{'BC}\in  \mathbb{R}^{(\hat{H}\hat{W}+\hat{H}\hat{W})\times C}$ and $S_t^{A\_BC} \in \mathbb{R}^{H^AW^A\times (\hat{H}\hat{W}+\hat{H}\hat{W})}$.  
As a result, we can reuse all the operations in the region correspondence module to undertake the group segmentation task. The above operations transform the network-like relations between individual images to a one-versus-rest relation, such that we can use the co-segmentation model to solve the group segmentation problem. 
We can repeat such operations to make predictions for each of the images in the group using the same parameters.

\subsection{Multi-Level Features for Co-Segmentation}
As the goal of co-segmentation is to segment common objects in two images, the similar regions in two images are important cues for prediction. We observe that such similarity may exist at multiple semantic levels. For example, if both the training and testing images contain the class \emph{car}, high-level features that correspond to object categories are useful for locating such a category. However, the testing images also contain novel classes that are never seen at training time. In this case, middle-level features that correspond to object parts would be useful, as they are more likely to be shared across classes. Based on such intuition, we design a model variant that exploits multi-level features in the encoder to undertake the co-segmentation task. We apply our cycle refinement module to features on different layers individually, and fuse their representations in the upsampling stage, as shown in Figure~\ref{fig:multi}. We adopt the feature fusion module proposed in \cite{yu2018learning} to fuse the representations from different levels. The multi-level version can also apply to the group-segmentation task with multiple input images. Multi-level features have proven useful in many vision tasks, such as semantic segmentation~\cite{ronneberger2015u} and object detection~\cite{ren2015faster}. Our experiments in Section~\ref{sec:ablation} show that using multi-level features to reason the common object region is helpful in the co-segmentation task.

\section{Experiment}

\begin{figure*}[t]
    \centering
    \includegraphics[width=0.95\linewidth]{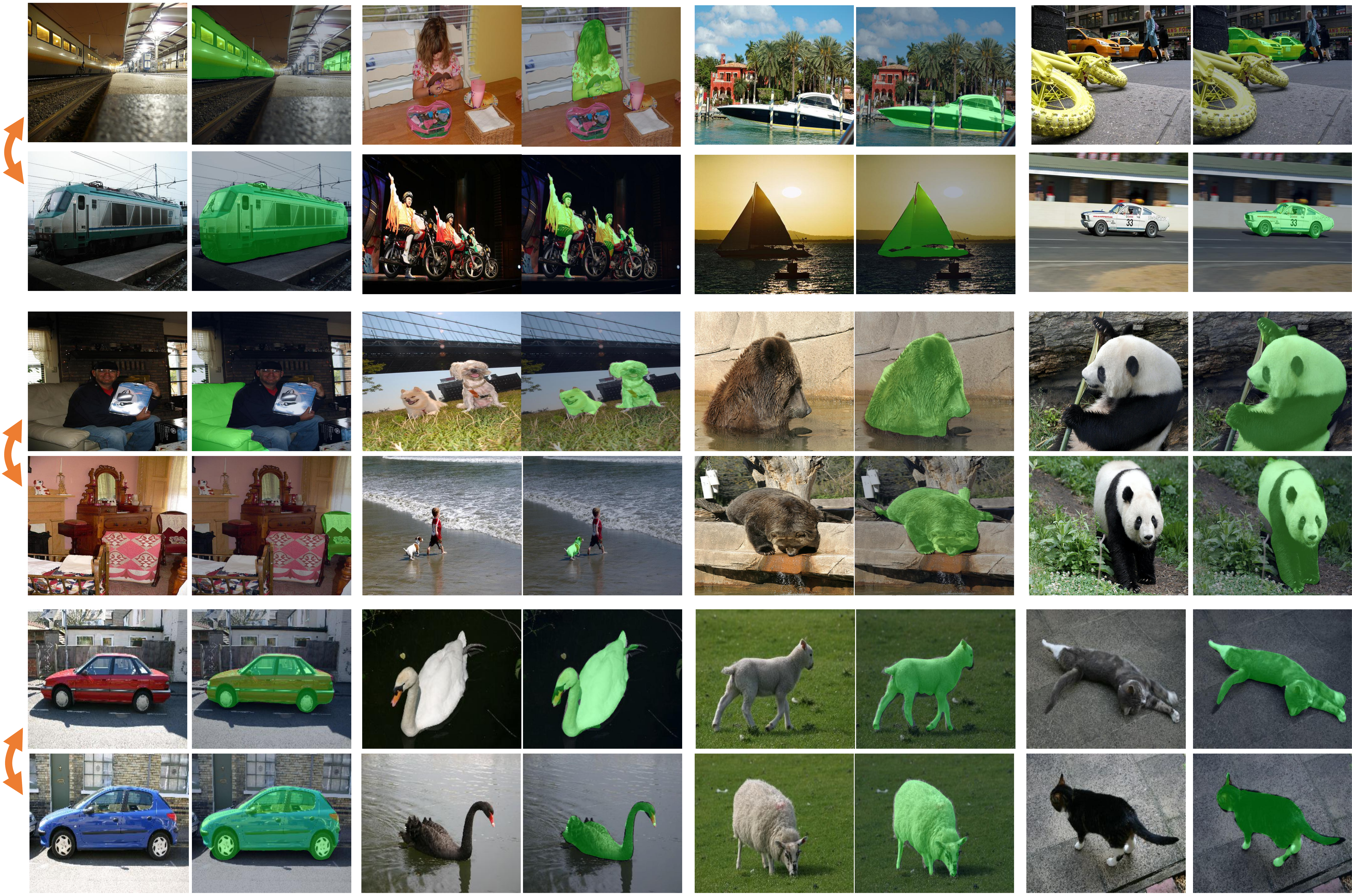}
    \caption{Qualitative results of our network on PASCAL VOC dataset, MSRC dataset, Internet dataset, and iCoseg dataset.}
    \label{fig:pascalvis}
\end{figure*}

\begin{figure*}[]
    \centering
    \includegraphics[width=0.9\linewidth]{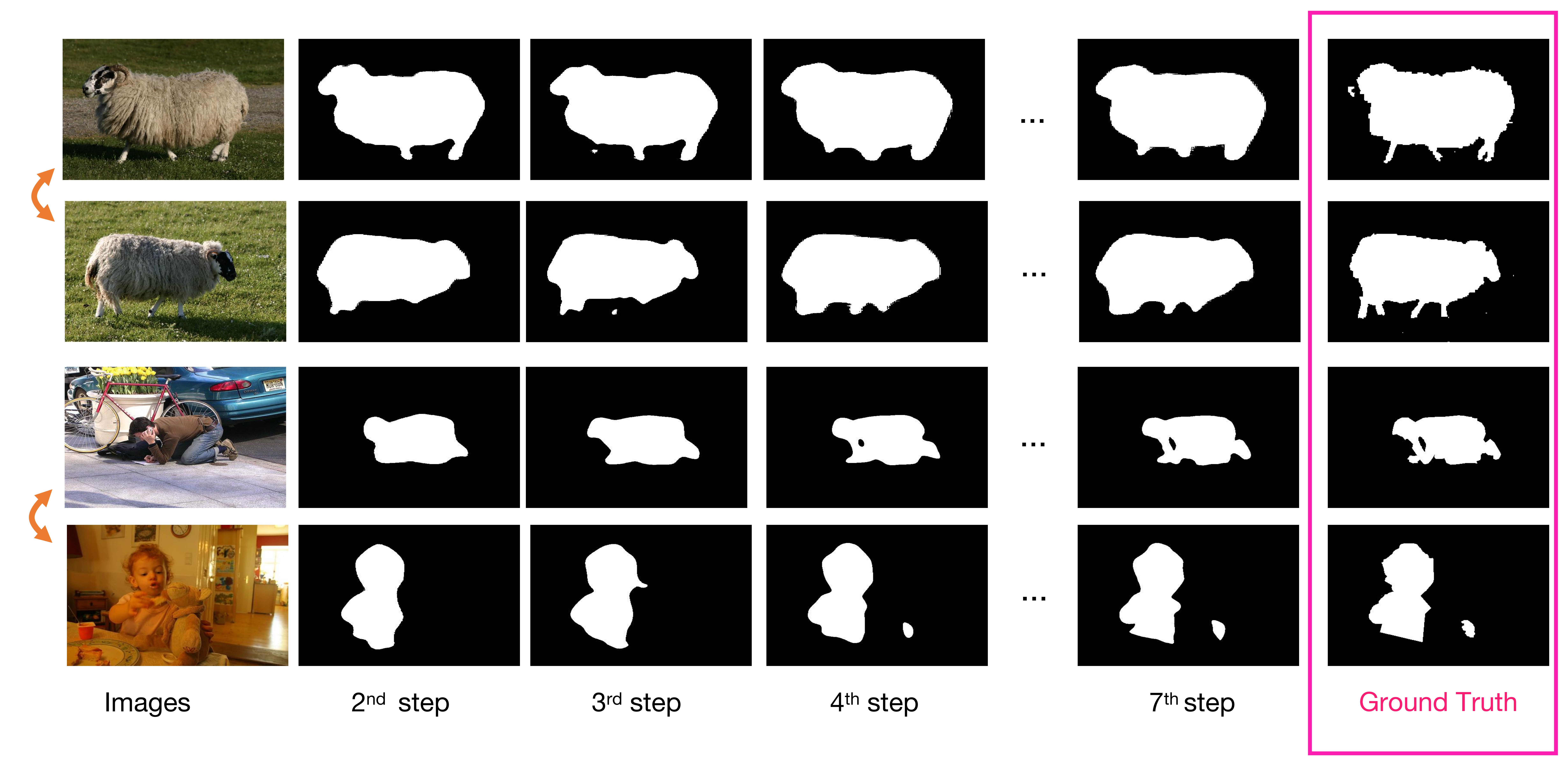}
    \caption{Predictions from different refinement steps. We decode the representations in each time step and the result shows that our proposed cycle refinement module can consistently improve the prediction. }
    \label{fig:Nvis}
\end{figure*}

\begin{figure}[]
    \centering
    \includegraphics[width=\linewidth]{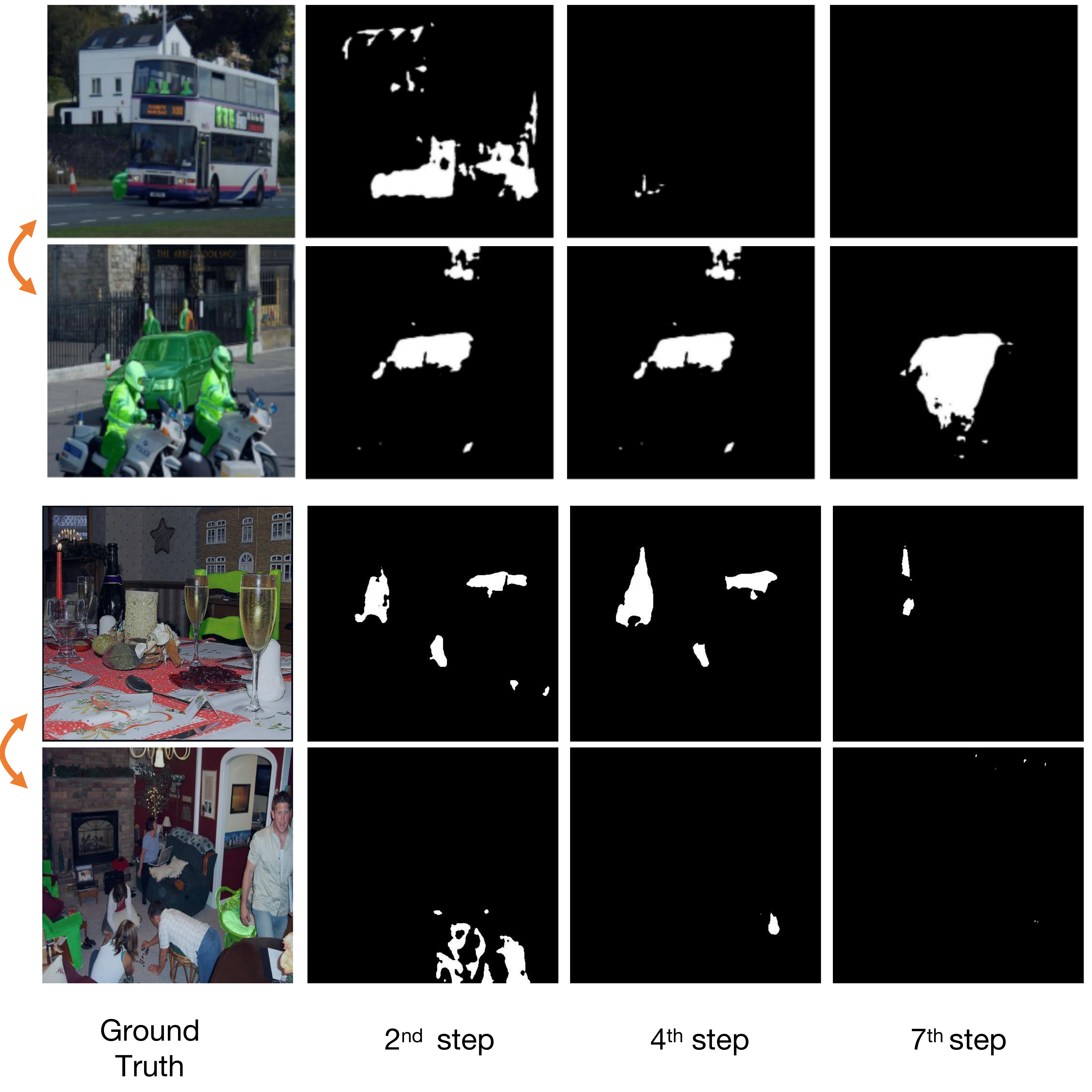}
    \caption{Failure cases. The ground truth is denoted with green masks in the RGB images. When the common objects demonstrate a large variance in appearance, the network may fail to establish meaningful connections between local regions, and the cycle refinement module can hardly further refine the predictions.  }
    \label{fig:failure}
\end{figure}

\subsection{Implementation Details.}
\textbf{Network.} 
The best performance of our network is obtained when we employ the ResNet34 \cite{he2016deep} as the encoder network and use multi-level features from the last three layers; the number of iterative steps is set as 7; the ROI region size is set as $2\times2$. If not further specified, we use them as the default setting in the experiments. The encoder is pretrained on ImageNet dataset~\cite{deng2009imagenet}. For the model version utilizing single-level features, we use the same decoder structure with~\cite{chen2018semantic} which contains five blocks of bilinear upsampling layers and convolutional layers.

\textbf{Training.} We use Adam algorithm~\cite{kingma2014adam} with the learning rate of 1e-5 to optimize the whole network in an end-to-end manner and set the weight decay as 0.0005.
For both training and evaluation, we resize every image from the datasets to the spatial size of 512${\times}$512. We set the batch size as 4 and train our network on the NVIDIA 1080TI GPU for 40K iterations. We use Lov${\acute{a}}$sz-Softmax proposed in \cite{berman2018lovasz} as the loss function, as we find it can yield slightly better performance than the pixel-wise cross-entropy loss. The overall loss function is  given as:
\begin{equation}
\begin{aligned}
& \mathcal{L}=\frac{1}{C} \sum_{c}  \bar{\Delta_{J_c}}(m(c)),  \\
&  m_{i}(c) =\left\{\begin{matrix}
1-p_{i}(c)\quad if \ \ c= y_{i}^{*}(c),\\ 
p_{i}(c) \quad  \quad \ \ otherwise,
\end{matrix}\right. 
\end{aligned}
\end{equation}
where ${C}$ denotes the number of classes, ${\bar{\Delta_{J_c}}\left (\cdot \right)}$ is the Lov${\acute{a}}$sz extension of the Jaccard Index, ${m(c)}$ is a vector of pixel errors for class ${c}$, ${y_i^{*}} {\in}\left \{ -1,1 \right \}$ is the ground truth label of pixel ${i}$ for class ${c}$,
and ${p_i(c)}$ ${\in} \left [0,1  \right ]$ is the predicted probability of pixel ${i}$ for class ${c}$. 
Please refer to \cite{berman2018lovasz} for more details about Lov${\acute{a}}$sz-Softmax loss function.

\begin{table*}[t]

\setcaptionmargin{7pt}
\begin{minipage}{\textwidth}
\begin{minipage}[t]{0.48\textwidth}
        \centering
          \begin{tabular}{ccc|ccc}
        \Xhline{1pt}
         $M_{cat}$ & $ M_{mul} $ & $ M_{cross} $ & $ M_{cycle} $    &  ${\mathcal{P} }$ (\%)  & ${\mathcal{J}}$ (\%)   \\
\hline\hline
 & & & & 77.2 & 58.6  \\
\checkmark & &  & & 81.3    & 62.9  \\
 & \checkmark &  &  &  83.6   & 64.2  \\
 & & \checkmark  &   &   87.1  &  68.2 \\
 & & \checkmark & \checkmark  & 91.8 & 70.4       \\

\Xhline{1pt}
\end{tabular}

\caption{Ablative analysis of the proposed model. ``${M_{cat}}$'', ``${M_{mul}}$'' and ``${M_{cross}}$'' denote three kinds of ways to exchange information between images. ``${M_{cycle}}$'' denotes our proposed  cycle refinement module. Our designs (``${M_{cross}}$'' and ``${M_{cycle}}$'') outperform the baselines significantly. Please refer to Section~\ref{sec:ablation} for analysis.}

\label{table:alb1}
\end{minipage}
\begin{minipage}[t]{0.48\textwidth}
        \centering
        \setlength{\tabcolsep}{5mm}{
            \begin{tabular}{cccc}
\Xhline{1pt}
\multicolumn{1}{c}{Num. of steps (${N}$)} & ${\mathcal{P}}$(\%) & ${\mathcal{J}}$(\%) & Time(s)  \\ \hline\hline
${N}$=${2}$&92.2 &71.9 &0.085 \\
${N}$=${3}$& 93.8 &73.1 & 0.088\\
${N}$=${4}$ & 94.2 &74.0 & 0.096 \\
${N}$=${5}$ & 94.8 &74.5 & 0.105\\
${N}$=${6}$ & 95.3 &75.2 &  0.109\\
${N}$=${7}$ &\textbf{95.8} &\textbf{75.4}  & \textbf{0.114} \\

\Xhline{1pt}
\end{tabular}}
\caption{Our network  with different refinement steps ${N}$ in the Cycle Refinement Module. As the refinement step increases, the performance grows consistently.}
\label{table:alb2}

\end{minipage}
\end{minipage}
\end{table*}

\begin{table*}[t]
\setcaptionmargin{3pt}
\setlength{\abovecaptionskip}{6pt}
    \begin{minipage}{\textwidth}
        \begin{minipage}[t]{0.50\textwidth}
        \centering
        \setlength{\tabcolsep}{7mm}{
        \begin{tabular}{ccc}
 
\Xhline{1pt}
\multicolumn{1}{c} {Region Size} & ${\mathcal{P}}$(\%) & ${\mathcal{J}}$(\%)  \\ \hline\hline
Raw   & 92.9& 73.1 \\
    $2 \times 2$     &\textbf{95.8} & \textbf{75.4} \\
    $3 \times 3$    & 94.0 & 74.1 \\
   $4 \times 4$     & 95.1 &74.8  \\
    $5 \times 5$    & 93.6 & 74.0 \\
    $6 \times 6$     & 94.8 &74.4 \\
    $7 \times 7$    &  94.6 & 74.3    \\

\Xhline{1pt}
\end{tabular}}
\caption{ Comparison of different ROI pooling sizes in the region correspondence module. The optimal result is obtained when the region size is $2 \times 2$.}
\label{table:alb3}
        
        \end{minipage}
        \begin{minipage}[t]{0.50\textwidth}
        \centering
        \setlength{\tabcolsep}{4mm}{
               \begin{tabular}{ccccc}
 
\Xhline{1pt}
\multicolumn{1}{c} {res\_5 \ } & {res\_4 \ } & {res\_3 \ } & ${\mathcal{P}}$(\%) & ${\mathcal{J}}$(\%)  \\ \hline\hline
\checkmark & &  & 91.8 & 70.4 \\
\checkmark &\checkmark &  & 93.5 & 72.5 \\
\checkmark & \checkmark & \checkmark & \textbf{95.8} & \textbf{75.4} \\
\hline
{vgg\_5 \ } & {vgg\_4 \ } & {vgg\_3 \ } & ${\mathcal{P}}$(\%) & ${\mathcal{J}}$(\%)   \\ \hline \hline
\checkmark & & & 91.2 & 69.8 \\
 \checkmark & \checkmark & & 93.1 & 71.8 \\
 \checkmark & \checkmark & \checkmark & \textbf{94.9} & \textbf{74.2} \\

\Xhline{1pt}
\end{tabular}
\caption{Performance of our method utilizing features from different stages in the encoder network. We conduct experiments with ResNet-34 and VGG16 as the network encoders. Using multi-level features can effectively boost network performance.}
    \label{table:alb4}}

        \end{minipage}
        
    \end{minipage}

\end{table*}

\begin{table*}[t]
\setlength{\abovecaptionskip}{5pt}
    \resizebox{\textwidth}{!}{
     \begin{tabular}{cccc|ccc|ccc|ccc}
 \Xhline{1pt} 
\multicolumn{1}{c}{\multirow{2}{*}{\textbf{Input Size}}} & \multicolumn{3}{c}{Strategy a)}  & \multicolumn{3}{c}{Strategy b)} & \multicolumn{3}{c}{Strategy c)}& \multicolumn{3}{c}{Strategy d)}  \\ 
\Xcline{2-13}{1pt}
\cline{2-13}
\multicolumn{1}{c}{}                        &  \multicolumn{1}{c}{Car  } & 
\multicolumn{1}{c}{Airplane  } & \multicolumn{1}{c}{Horse } & 
\multicolumn{1}{c}{Car  } & 
\multicolumn{1}{c}{Airplane  } & \multicolumn{1}{c}{Horse } & 
\multicolumn{1}{c}{Car  } &
\multicolumn{1}{c}{Airplane  } & \multicolumn{1}{c}{Horse } & 
\multicolumn{1}{c}{Car } & 
\multicolumn{1}{c}{Airplane } & 
\multicolumn{1}{c}{Horse }   \\
\Xhline{1pt}
 ${k=2}$& 85.8 &77.2 &74.1 & 84.4& 78.3 & 73.9  &85.9 & 77.3  &74.0   & 85.7 & 77.3 &74.1  \\
 ${k=3}$ &86.6 &78.1 &75.0 & 84.8& 78.8 & 74.1  &86.6 & 78.1  &75.0  & 86.8 &78.5 &75.1 \\
 ${k=4}$&- &- &- &  85.6 & 78.9 & 75.1   &86.9 &78.6 &75.4  &\textbf{87.0} &78.8 & \textbf{75.5} \\
 ${k=5}$&- &- &- &  86.5 &  \textbf{79.3} & 75.4  &86.9 &78.6  &\textbf{75.5}  & \textbf{87.0} & 78.8 &\textbf{75.5}\\
 ${k=6}$&- &- &- & 86.4 & 79.0  &75.4 & 86.9 & 78.7 &75.4  & 86.9 & 78.7 & \textbf{75.5} \\
 ${k=7}$&- &- &- & 86.3& 78.6 &75.2  & 86.7 & 78.7  & 75.4 & 86.8 & 78.7 &   75.4   \\
 ${k=8}$&- &- &- &86.1& 78.9  & 75.3 & 86.7 & 78.7  & \textbf{75.5} & 86.8 & 78.7 &   \textbf{75.5}   \\
 \Xhline{1pt} 
\end{tabular}
}

    \caption{The results (Jaccard) of group segmentation with different sampling strategies and different numbers of input images on  Internet dataset. Our network can handle more than two images at a time, which generates better results than our network with paired input images. Please refer to Section~\ref{sec:ablation} for the description and analysis of different strategies. }
    \label{tab:compare}
\end{table*}

\begin{table*}[t]
\centering

\setlength{\abovecaptionskip}{0pt}
	\begin{minipage}{0.95\textwidth}
	\centering
		\begin{minipage}[t]{0.46\textwidth}
			\centering
			\begin{subtable}{\textwidth}
			\centering
			\begin{tabular}{cccc}
					\toprule[0.9pt]
\multicolumn{1}{c}{\textbf{VOC2012(pair)}  } &  \textbf{Training Set} & \ ${\mathcal{P}}$(\%) & \ ${\mathcal{J}}$(\%)  \\ 
\Xhline{0.9pt}
CA\cite{chen2018semantic} & VOC2012 & -&    59.2             \\ 
FCA\cite{chen2018semantic} & VOC2012 &-&    59.4            \\
CSA\cite{chen2018semantic} & VOC2012 &  -&    59.8             \\

DOCS\cite{li2018deep} & VOC2012  &  94.2&    64.5             \\ 
\Xhline{0.9pt}
\textbf{CycleSegNet}(Ours) & VOC2012 &  \textbf{95.8}&    \textbf{75.4} \\ 
\Xhline{0.9pt}					

				\end{tabular}

		\setcaptionmargin{3pt}
	    \caption{Pairwise co-segmentation results on the \textbf{PASCAL VOC2012} dataset.}
		\label{table:pascal}
		\end{subtable}
		\end{minipage}
		\begin{minipage}[t]{0.46\textwidth}
			\centering
			\begin{subtable}{\textwidth}
			\centering
			\begin{tabular}{cccc}
					\toprule[0.9pt]
\multicolumn{1}{c}{\textbf{VOC2010 (group)}  } & \textbf{Training Set}  & \ ${\mathcal{P}}$(\%) & \ ${\mathcal{J}}$(\%)  \\ 
\Xhline{0.9pt}

Quan~\textit{et al.}\cite{quan2016object} &-  & 89.0  & 52.0  \\
Jerripothula~\textit{et al.}\cite{jerripothula2016image} & - & 85.2 & 45.0 \\
Jerripothula~\textit{et al.}\cite{jerripothula2017object} &-  & 80.1 & 40.0\\
Li~\textit{et al.}\cite{li2019group} & COCO &  94.1&    63.0            \\ 
Zhang~\textit{et al.}\cite{zhang2019deep} & COCO  &  94.9 &    71.0             \\
\Xhline{0.9pt}
\textbf{CycleSegNet}(Ours) & COCO &  \textbf{96.8}&    \textbf{73.6} \\ 
\Xhline{0.9pt}					

				\end{tabular}


				
		\setcaptionmargin{3pt}
	    \caption{Group segmentation results on \textbf{PASCAL VOC2010} dataset.}
		\label{table:pascal2}
		\end{subtable}
		\end{minipage}
		\begin{minipage}[t]{0.70\textwidth}
			\vspace{5pt}
\centering
			\begin{subtable}{\textwidth}
			    \centering
			    \begin{tabular}{lccc|cc}
						\toprule[1pt]
\multicolumn{1}{c}{\multirow{2}{*}{\textbf{MSRC}}}&
\multicolumn{1}{c}{\multirow{2}{*}{\textbf{Training Set}}}
& \multicolumn{2}{c}{group}  & \multicolumn{2}{c}{pair}  \\ \Xcline{3-6}{0.5pt}%
\cline{3-6}
\multicolumn{1}{c}{}                        & 
\multicolumn{1}{c}{}                        & \multicolumn{1}{c}{${\mathcal{P}}$(\%)} & \multicolumn{1}{c}{${\mathcal{J}}$(\%)} & \multicolumn{1}{c}{${\mathcal{P}}$(\%))} & \multicolumn{1}{c}{${\mathcal{J}}$(\%)} \\
\Xhline{1pt}
Vicente~\textit{et al.}\cite{vicente2011object}& - & 90.2 & 70.6 & - & -\\
Wang~\textit{et al.}\cite{wang2013image} & - & 92.2 & - & - & -\\
Rubinstein~\textit{et al.}\cite{rubinstein2013unsupervised} & - & 92.2 & 74.7 & - & -\\
Faktor~\textit{et al.}\cite{faktor2013co} & - & 92.0 & 77.0 & - & -\\
Chen~\textit{et al.}\cite{chen2018semantic} &VOC2012 &-  &  73.9 &  96.6&    76.5  \\ 
Li~\textit{et al.}\cite{li2018deep} &VOC2012 &95.4  &82.9    &- &-              \\

\Xhline{1pt}

\textbf{CycleSegNet}(Ours) &VOC2012 & \textbf{97.9}&    \textbf{87.2}   & \textbf{97.3} &  \textbf{86.8}          \\ 

\hline \hline\Xhline{0.5pt}
Zhang~\textit{et al.}\cite{zhang2019deep} &COCO &  95.2&    81.9        &- &-     \\ 

\Xhline{1pt}
\textbf{CycleSegNet}(Ours)&COCO &  \textbf{97.6}&    \textbf{89.6}   & \textbf{97.2} &  \textbf{88.2}          \\ 
\Xhline{1pt}

					\end{tabular}
				\setcaptionmargin{3pt}
				\caption{Results on the \textbf{MSRC} dataset.}
				\label{table:msrc}
			\end{subtable}

		\end{minipage}

		\begin{minipage}[t]{0.95\textwidth}
		    \vspace{5pt}
		    \centering
		    \begin{subtable}{\textwidth}
		    \begin{tabular}{lccccc|cccc}
		            \toprule[1pt]
\multicolumn{1}{c}{\multirow{2}{*}{\textbf{Internet}}} & 
\multicolumn{1}{c}{\multirow{2}{*}{\textbf{Training Set}}} &
\multicolumn{4}{c}{group}  & \multicolumn{4}{c}{pair}  \\ \Xcline{3-10}{0.5pt}
\multicolumn{1}{c}{}                        & 
\multicolumn{1}{c}{}                        & \multicolumn{1}{c}{Airplane} & \multicolumn{1}{c}{Car} & \multicolumn{1}{c}{Horse} & \multicolumn{1}{c}{Avg.${\mathcal{J}}$(\%)} &
\multicolumn{1}{c}{Airplane} & \multicolumn{1}{c}{Car} & \multicolumn{1}{c}{Horse} & \multicolumn{1}{c}{Avg.${\mathcal{J}}$(\%)}  \\\Xhline{1pt}

Rubinstein~\textit{et al.}\cite{rubinstein2013unsupervised} & - & 55.8 & 64.4 & 51.6& 57.3 &- &-&- &- \\

Quan~\textit{et al.}\cite{quan2016object} & - & 56.3 & 66.8 & 58.1  & 60.4  &- &-&- &- \\
Jerripothula~\textit{et al.}\cite{jerripothula2016image} & - & 61.0 & 71.0 & 60.0 & 64.0 &- &-&- &- \\

Chen~\textit{et al.}\cite{chen2019show} &- & 65.0 & 82.0 & 63.0 & 70.0 &- &-&- &- \\

Yuan~\textit{et al.}\cite{yuan2017deep} & VOC2012 & 66.0 & 72.0 & 65.0 & 67.7 &- &-&- &- \\
 
Li~\textit{et al.}\cite{li2018deep} &VOC2012 & 65.4 & 82.8 & 69.4 & 72.6 &- &-&- &- \\
Chen~\textit{et al.}\cite{chen2018semantic}&VOC2012 &- &-&- &-  & 71.4 &    79.9 & 68.0  & 73.1            \\ 

\Xhline{1pt}
\textbf{CycleSegNet}(Ours)&VOC2012  &  \textbf{78.8} &    \textbf{87.0}  & \textbf{75.5} &    \textbf{80.4}  & \textbf{77.2} &  \textbf{85.8}   & \textbf{74.1}  &   \textbf{79.0}           \\ 
\hline \hline
\Xhline{1pt}

Zhang~\textit{et al.}\cite{zhang2019deep}&COCO  & 69.6&  82.5 & 70.2 & 74.1 &- &-&- &-\\

Li~\textit{et al.}\cite{li2019group}&COCO  &83.0&  \textbf{93.0} & 76.0 & 84.0 &- &-&- &-\\

\Xhline{1pt}
\textbf{CycleSegNet}(Ours) &COCO &  \textbf{84.1} &    91.6  & \textbf{82.9} &    \textbf{86.2}  &   \textbf{83.2}   & \textbf{88.5}  &  \textbf{80.3}  & \textbf{84.0}           \\ 
\Xhline{1pt}
		            \end{tabular}

		        \caption{Results on \textbf{Internet} dataset.}
		        \label{table:internet}
		    \end{subtable}
		\end{minipage}
		
		\begin{minipage}[t]{0.95\textwidth}
		    \vspace{5pt}
			\centering
			\begin{subtable}{\textwidth}
			
			\resizebox{\textwidth}{!}{
				\begin{tabular}{lcccccccccc}
                \toprule[1pt] 

\multicolumn{1}{c}{\multirow{1}{*}{\textbf{iCoseg} (pair)}} 
                      & 
                      \multicolumn{1}{c}{\multirow{1}{*}{\textbf{Training Set}}} 
                      & \multicolumn{1}{c}{bear2} & \multicolumn{1}{c}{brownbear} & \multicolumn{1}{c}{cheetah} & \multicolumn{1}{c}{elephant} &
\multicolumn{1}{c}{helicopter} & \multicolumn{1}{c}{hotballoon} & \multicolumn{1}{c}{panda1} &\multicolumn{1}{c}{panda2} & Avg.${\mathcal{J}}$(\%) \\\Xhline{1pt}
Li~\textit{et al.}\cite{li2018deep}&VOC2012  & 88.3 & 92.0 & 68.8 & 84.6 & 79.0 & 91.7 & 82.6 & 86.7 & 84.2   \\ 
Chen~\textit{et al.}\cite{chen2018semantic} &VOC2012 & 88.3 & 91.5 & 71.3 & 84.4 & 76.5 & 94.0 & 91.8 & 90.3 & 86.0   \\ 

\Xhline{1pt}
\textbf{CycleSegNet}(Ours)&VOC2012 &  \textbf{92.1} &  \textbf{94.8}  & \textbf{84.5} &   \textbf{90.2} &  \textbf{80.2} &    \textbf{95.8}  & \textbf{94.5}  & \textbf{94.1} & \textbf{90.8}        \\ 
\Xhline{1pt}

\end{tabular}
		}
		\caption{ Pairwise co-segmentation results on \textbf{iCoseg} dataset.}
		\label{table:icoseg}
		\end{subtable}
		
		\end{minipage}
		
		\begin{minipage}[t]{0.95\textwidth}
		    \vspace{5pt}
			\centering
			\begin{subtable}{\textwidth}
			\resizebox{\textwidth}{!}{
				\begin{tabular}{lcccccccccc}
                \toprule[1pt] 
\multicolumn{1}{c}{\multirow{1}{*}{\textbf{iCoseg} (group)}}  
                       & \multicolumn{1}{c}{\multirow{1}{*}{\textbf{Training Set}}}  
                       &  \multicolumn{1}{c}{bear2} & \multicolumn{1}{c}{brownbear} & \multicolumn{1}{c}{cheetah} & \multicolumn{1}{c}{elephant} &
\multicolumn{1}{c}{helicopter} & \multicolumn{1}{c}{hotballoon} & \multicolumn{1}{c}{panda1} &\multicolumn{1}{c}{panda2} & Avg.${\mathcal{J}}$(\%) \\\Xhline{1pt}

Rubinstein~\textit{et al.}\cite{rubinstein2013unsupervised} & - & 65.3 & 73.6 & 69.7 & 68.8 & 80.3& 65.7 & 75.9 & 62.5 & 70.2 \\
Jerripothula~\textit{et al.}\cite{jerripothula2014automatic} & - & 70.1 &66.2& 75.4 & 73.5 & 76.6 & 76.3 & 80.6 & 71.8 & 73.8 \\ 
Faktor~\textit{et al.}\cite{faktor2013co} & - & 72.0 & 66.2 &. 75.4& 73.5 & 76.6 & 76.3 & 80.6 & 71.8 & 73.8 \\
Jerripothula~\textit{et al.}\cite{jerripothula2016image} &- &67.5 & 72.5 & 78.0 & 79.9 & 80.0 & 80.2 & 72.2 & 61.4 & 70.4     \\ 
Zhang~\textit{et al.}\cite{zhang2019deep}&COCO & 91.1 & 89.6 & 88.6 & 90.0 & 76.4 & 94.2 & 90.4 & 87.5 & 89.2   \\ 

\Xhline{1pt}
 
\textbf{CycleSegNet}(Ours) &COCO &  \textbf{92.8} &    \textbf{94.1}  & \textbf{89.1} &    \textbf{91.6} &  \textbf{85.2} &    \textbf{95.1}  & \textbf{94.8}  & \textbf{94.1}  & \textbf{92.1}          \\ 
\Xhline{1pt}

\end{tabular}
		}
		\caption{  Group segmentation results on \textbf{iCoseg} dataset.}
		\label{table:icoseg2}
		\end{subtable}
		
		\end{minipage}
	
    \caption{ Comparison with the state-of-the-art performance of pairwise co-segmentation and group segmentation tasks on four benchmark datasets: \textbf{PASCAL VOC} dataset, \textbf{MSRC} dataset, \textbf{Internet} dataset and \textbf{iCoseg} dataset. Our model achieves new state-of-the-art performance on all datasets. }
	\label{table:soa}	
	\end{minipage}
	\vskip -1em
	
\end{table*}

\subsection{Datasets and Evaluation Metric}
As previous models use different training datasets in this task, we report the performance of our model with different training datasets for a fair comparison. Specifically,
in~\cite{li2018deep,chen2018semantic}, the training set is constructed based on the training set of PASCAL VOC2012 dataset while in~\cite{li2019group,zhang2019deep}, the training set is based on COCO dataset~\cite{lin2014microsoft}.
We evaluate our proposed method and compare it with existing methods on four widely-used benchmark datasets for object co-segmentation, including PASCAL VOC dataset\cite{everingham2015pascal}, MSRC dataset\cite{shotton2006textonboost}, Internet dataset\cite{rubinstein2013unsupervised}, and iCoseg dataset\cite{batra2010icoseg}. 

\textbf{PASCAL VOC.}
 PASCAL VOC2012 dataset contains 20 foreground object classes and one background class and it has 1,464 training images and 1,449 validation images in total.
Following~\cite{li2018deep,chen2018semantic,wang2019zero}, we split the original validation set into a validation set (724 images)  and a test set (725 images) for the co-segmentation task.  For PASCAL VOC 2010, we follow the setting in ~\cite{li2019group,zhang2019deep}, where a total of 1,037 images of 20 objects classes from PASCAL VOC 2010 dataset are used for evaluation.

\textbf{MSRC.}
Following ~\cite{li2018deep,zhang2019deep}, we use the same subset of MSRC dataset. This subset includes 7 classes: bird, car, cat, cow, dog, plane, and sheep. Each class contains 10 images.

\textbf{Internet.}
 Internet dataset contains images of three categories including airplane, car, and horse. Following previous works~\cite{hsu2018co,chen2018semantic}, we use the same subset of the Internet dataset where each class has 100 images.

\textbf{iCoseg.}
iCoseg dataset consists of 643 images from 38 categories. Large variances of viewpoints and deformations are present in this dataset. A subset that contains 8 classes is used to evaluate the generalizability of our proposed method.

We evaluate the performance of co-segmentation models using two common evaluation metrics: \textit{Precision} (${\mathcal{P}}$) and \textit{Jaccard Index} (${\mathcal{J}}$).
We report the performance under both metrics on these four co-segmentation benchmark datasets in our experiments. All the  experiments for analysis are trained and evaluated on PASCAL VOC 2012 dataset.

\subsection{Analysis}
\label{sec:ablation}
In this part, we investigate the effectiveness of each component in our proposed CycleSegNet by ablative analysis.
We first create a baseline model which is a foreground prediction network where two branches do not exchange any information. Then we incorporate another two baseline methods $M_{cat}$ and $M_{mul}$ that exchange information in the form of global representations. Concretely, $M_{cat}$ is to upsample the global vector and fuse it with the feature maps by concatenation, while $M_{mul}$ is to fuse by multiplication. 
Then we add each of our proposed modules to the baseline model one by one. All the models are tested with the single-level feature encoder. The results are shown in Table~\ref{table:alb1}. As we can see, both our proposed modules are very effective in improving the performance over the baseline networks. 
Specifically, our region correspondence module ($M_{cross}$) significantly outperforms both baseline methods $M_{cat}$ and $M_{mul}$ by 5.8\% and 3.5\% in terms of Precision and 5.3\% and 4.0\% in terms of Jaccard, which shows that our design that exchanges information between local regions is more effective than previous methods using global representations.
The cycle refinement module further boosts the Precision and Jaccard score by 4.7\% and 2.2\%, respectively.

\textbf{Number of the refinement step.} We report the performance of our network under different refinement steps ${N}$ in Table~\ref{table:alb2}. We also compare the average time to inference one pair of images.
As we can see from the table, when the number of the refinement steps increases, the performance in terms of both Precision and Jaccard increases consistently while the increment in inference time is marginal. Particularly, after 7 steps of refinement, the network can improve the initial predictions by 3.6\% and 3.5\% in terms of Precision and Jaccard, respectively.

\textbf{The size of the regional representations.} 
In the region correspondence module, we use the ROI pooling to get the regional representations of images.
We next investigate how the region size influences the performance in Table~\ref{table:alb3}. As we can see, the best performance is observed when ROI pooling outputs a 2 ${\times}$ 2 feature map. Directly using the original representations without ROI pooling does not yield better performance than regional representations. A possible reason is that the feature points in the original feature representations have a relatively small effective field-of-view and thus lack the expressive power of abstract and semantic concepts.
Using the ROI pooling can enlarge the field-of-view and provide context information in local regions, which is helpful in computing the correlation scores between regions for information exchange.

\textbf{Multi-level features.}
We compare the performance of our network using features from different layers in the encoder network in Table~\ref{table:alb4}. 
As can be observed, multi-level features can  boost the performance of our network with both the VGG16 backbone and the ResNet34 backbone.
Specifically, using the features from the last three layers is better than using the features from the last two layers. This indicates that middle-level features are helpful and can provide important cues in the co-segmentation task.

\textbf{Group Segmentation.}
We present a detailed analysis about group segmentation in this part. The group segmentation aims to predict the common objects in a group of images.
We apply different CNN based solutions in previous works to our network and then compare their performance.
Since the networks in most previous works for co-segmentation can only handle paired inputs, they often adopt various sampling strategies and fusion methods to undertake the group segmentation tasks, which can be applied to our network as well. An advantage of our network is that our network can handle more than two images at a time, as is described earlier, so the number of input images is flexible. However, directly sending all images in the group to our network is unrealistic due to the limited GPU memory. We therefore sample a small group of $k$ images at a time and send them to our network for predictions. Then, we fuse all the predictions of an image as the final predicted mask. Suppose that there are $N$ images to be segmented in a group segmentation task and our network handles a tuple of $k$ images at a time. We compare the following strategies to undertake group segmentation:\\
\textbf{a)} Following \cite{banerjee2019cosegnet}, we sample all possible $k$-element tuples to make predictions and average all the corresponding confidence maps of an image as the final prediction. This method can make use of all images in the group to make predictions for each image.\\
\textbf{b)} Following \cite{zhang2019deep}, we randomly divide all images into $N/k$ small groups and each group has $k$ images. In this case, each image is only tested once and the whole evaluation process is quick.\\
\textbf{c)} Following \cite{li2018deep}, to predict the mask of an image, we randomly sample 5 groups of images from the other $N-1$ images, where each group has $k-1$ images. Then, the target image joins each of these groups to make a prediction with our network, and we average $5$ predictions to generate the final mask. The process is repeated for all  $N$ images.\\
\textbf{d)} Following \cite{wang2019zero}, to predict the mask of an image, we uniformly split the other ${N-1}$ images into ${T}$ groups, where ${T=(N-1)/(k-1) }$ and each group has $k-1$ images. Then, the target image joins each of these groups to make a prediction with our network, and we average $T$ predictions to generate the final mask. The process is repeated for all $N$ images.\\
Besides these solutions, there are also some previous works using graphs~\cite{quan2016object,han2017robust} or recurrent models~\cite{wang2019zero} to undertake group segmentation, which, however, can not be applied to our network directly. Therefore, we compare the four solutions above with our network. We conduct experiments on the subset of the Internet dataset, where each class has 100 images. 
For the solution \textbf{a}, there would be too many possible tuples for testing  when $k$ is large. Therefore, we only test the cases where $k=2$ and $k=3$.
The result is shown in Table~\ref{tab:compare}. 
As we can see, predicting with more than two input images is consistently better than predicting with paired input images with all four strategies, which shows the superiority of our network design for group segmentation. The best tuple size is 4 or 5, while further increasing the tuple size can not boost the performance.

\textbf{Visualization.}
We present the visualization of our network predictions to qualitatively evaluate our design in Figure~\ref{fig:pascalvis}. As we can see, our model can well segment the common objects from the complex scenes in two images. To further observe how our cycle refinement module improves the predictions, we visualize the predictions in each refinement step, as shown in Figure~\ref{fig:Nvis}. We can see from the result that the predicted masks become more accurate with more refinement steps. We also present some failure cases in Figure~\ref{fig:failure}. As is shown, when the objects in different images demonstrate a large appearance variance, such as huge differences in object sizes and poses, the network fails to establish meaningful correspondence between local regions for predictions, and the CRM can not gradually improve the results.

\subsection{Comparison with the State-of-the-Arts}
Finally, we compare our proposed network with the state-of-the-art (SOTA) methods on four popular benchmark datasets: PASCAL VOC dataset, MSRC dataset, Internet dataset, and iCoseg dataset.
We report the model performance for group segmentation as well as the pairwise co-segmentation. 
The results are shown in Table~\ref{table:soa}.
As can be seen, our method outperforms all previous methods on different benchmarks with  significant margins.
In particular, on the challenging PASCAL VOC2012 dataset, the performance of our method outperforms the SOTA pairwise co-segmentation result by \textbf{10.9\%} in terms of Jaccard (Table~\ref{table:soa} (a)); 
on the MSRC dataset, when the models are trained with PASCAL VOC 2012 dataset, our performance of group segmentation and pairwise co-segmentation outperform the SOTA result by \textbf{4.3\%} and \textbf{10.3\%}, respectively, in terms of Jaccard (Table~\ref{table:soa} (c)); 
on Internet dataset, the performance of our model trained with PASCAL VOC 2012 dataset on the group segmentation task and pairwise co-segmentation task outperforms the SOTA result by \textbf{7.8\%} and \textbf{5.9\%}, respectively, in terms of Jaccard (Table~\ref{table:soa}(d));
on iCoseg dataset, our pairwise co-segmentation performance outperforms the SOTA result by  \textbf{4.8\%}, in terms of Jaccard  (Table~\ref{table:soa}(e));

\section{Conclusions}
In this paper, we propose a novel and effective approach for the image co-segmentation task.
The proposed region correspondence module directly exchanges information between local regions from different images, which demonstrates advantages over the baseline methods that transfer global image representations. The cycle refinement module that employs ConvLSTMs to progressively exchange information between images and update image representations can consistently improve the network predictions. Our algorithm can handle the input with paired images as well as a group of images with the same network and parameters. The multi-level feature encoder can further boost the network performance effectively.
Experiment results on four object co-segmentation datasets --- PASCAL VOC dataset, MSRC dataset, Internet dataset, and iCoseg dataset demonstrate that our proposed method significantly outperforms the existing methods and achieves new state-of-the-art performance.

\ifCLASSOPTIONcompsoc
  \section*{Acknowledgments}
\else
  \section*{Acknowledgment}
\fi

This research is supported by the National Research Foundation, Singapore under its AI Singapore Programme (AISG Award No: AISG-RP-2018-003), the MOE Tier-1 research grants: RG28/18 (S), RG22/19 (S) and RG95/20, and the National Natural Science Foundation of China (No. 61772530). 
This work is also supported by National Natural Science Foundation of China (NSFC) 61876208, Key-Area Research and Development Program of Guangdong Province 2018B010108002, and Central Universities of China under Grant D2192860.

{\small
\bibliographystyle{splncs04}
\bibliography{egbib}

\begin{thebibliography}{10}
\providecommand{\url}[1]{\texttt{#1}}
\providecommand{\urlprefix}{URL }
\providecommand{\doi}[1]{https://doi.org/#1}

\bibitem{banerjee2019cosegnet}
Banerjee, S., Hati, A., Chaudhuri, S., Velmurugan, R.: Cosegnet: Image
  co-segmentation using a conditional siamese convolutional network. In: Proc.
  28th Int. Joint Conf. Artif. Intell. pp. 673--679 (2019)

\bibitem{batra2010icoseg}
Batra, D., Kowdle, A., Parikh, D., Luo, J., Chen, T.: icoseg: Interactive
  co-segmentation with intelligent scribble guidance. In: 2010 IEEE Computer
  Society Conference on Computer Vision and Pattern Recognition. pp.
  3169--3176. IEEE (2010)

\bibitem{berman2018lovasz}
Berman, M., Rannen~Triki, A., Blaschko, M.B.: The lov{\'a}sz-softmax loss: a
  tractable surrogate for the optimization of the intersection-over-union
  measure in neural networks. In: Proceedings of the IEEE Conference on
  Computer Vision and Pattern Recognition. pp. 4413--4421 (2018)

\bibitem{chang2011co}
Chang, K.Y., Liu, T.L., Lai, S.H.: From co-saliency to co-segmentation: An
  efficient and fully unsupervised energy minimization model. In: CVPR 2011.
  pp. 2129--2136. IEEE (2011)

\bibitem{chen2018semantic}
Chen, H., Huang, Y., Nakayama, H.: Semantic aware attention based deep object
  co-segmentation. In: Asian Conference on Computer Vision. pp. 435--450.
  Springer (2018)

\bibitem{chen2015co}
Chen, H.Y., Lin, Y.Y., Chen, B.Y.: Co-segmentation guided hough transform for
  robust feature matching. IEEE transactions on pattern analysis and machine
  intelligence  \textbf{37}(12),  2388--2401 (2015)

\bibitem{chen2020compositional}
Chen, X., Zhang, C., Lin, G., Han, J.: Compositional prototype network with
  multi-view comparision for few-shot point cloud semantic segmentation. arXiv
  preprint arXiv:2012.14255  (2020)

\bibitem{chen2019show}
Chen, Y.C., Lin, Y.Y., Yang, M.H., Huang, J.B.: Show, match and segment: Joint
  learning of semantic matching and object co-segmentation. arXiv preprint
  arXiv:1906.05857  (2019)

\bibitem{deng2009imagenet}
Deng, J., Dong, W., Socher, R., Li, L.J., Li, K., Fei-Fei, L.: Imagenet: A
  large-scale hierarchical image database. In: 2009 IEEE conference on computer
  vision and pattern recognition. pp. 248--255. Ieee (2009)

\bibitem{everingham2015pascal}
Everingham, M., Eslami, S.A., Van~Gool, L., Williams, C.K., Winn, J.,
  Zisserman, A.: The pascal visual object classes challenge: A retrospective.
  International journal of computer vision  \textbf{111}(1),  98--136 (2015)

\bibitem{faktor2013co}
Faktor, A., Irani, M.: Co-segmentation by composition. In: Proceedings of the
  IEEE International Conference on Computer Vision. pp. 1297--1304 (2013)

\bibitem{fan2021re}
Fan, D.P., Li, T., Lin, Z., Ji, G.P., Zhang, D., Cheng, M.M., Fu, H., Shen, J.:
  Re-thinking co-salient object detection. IEEE Transactions on Pattern
  Analysis and Machine Intelligence  (2021)

\bibitem{fu2019stacked}
Fu, J., Liu, J., Wang, Y., Zhou, J., Wang, C., Lu, H.: Stacked deconvolutional
  network for semantic segmentation. IEEE Transactions on Image Processing
  (2019)

\bibitem{guo2019degraded}
Guo, D., Pei, Y., Zheng, K., Yu, H., Lu, Y., Wang, S.: Degraded image semantic
  segmentation with dense-gram networks. IEEE Transactions on Image Processing
  \textbf{29},  782--795 (2019)

\bibitem{han2017unified}
Han, J., Cheng, G., Li, Z., Zhang, D.: A unified metric learning-based
  framework for co-saliency detection. IEEE Transactions on Circuits and
  Systems for Video Technology  \textbf{28}(10),  2473--2483 (2017)

\bibitem{han2017robust}
Han, J., Quan, R., Zhang, D., Nie, F.: Robust object co-segmentation using
  background prior. IEEE Transactions on Image Processing  \textbf{27}(4),
  1639--1651 (2017)

\bibitem{he2017mask}
He, K., Gkioxari, G., Doll{\'a}r, P., Girshick, R.: Mask r-cnn. In: Proceedings
  of the IEEE international conference on computer vision. pp. 2961--2969
  (2017)

\bibitem{he2016deep}
He, K., Zhang, X., Ren, S., Sun, J.: Deep residual learning for image
  recognition. In: Proceedings of the IEEE conference on computer vision and
  pattern recognition. pp. 770--778 (2016)

\bibitem{hsu2018co}
Hsu, K.J., Lin, Y.Y., Chuang, Y.Y.: Co-attention cnns for unsupervised object
  co-segmentation. In: IJCAI. pp. 748--756 (2018)

\bibitem{hu2019attention}
Hu, T., Yang, P., Zhang, C., Yu, G., Mu, Y., Snoek, C.G.: Attention-based
  multi-context guiding for few-shot semantic segmentation. In: Proceedings of
  the AAAI conference on artificial intelligence. vol.~33, pp. 8441--8448
  (2019)

\bibitem{jerripothula2017object}
Jerripothula, K.R., Cai, J., Lu, J., Yuan, J.: Object co-skeletonization with
  co-segmentation. In: 2017 IEEE Conference on Computer Vision and Pattern
  Recognition (CVPR). pp. 3881--3889. IEEE (2017)

\bibitem{jerripothula2014automatic}
Jerripothula, K.R., Cai, J., Meng, F., Yuan, J.: Automatic image
  co-segmentation using geometric mean saliency. In: 2014 IEEE International
  Conference on Image Processing (ICIP). pp. 3277--3281. IEEE (2014)

\bibitem{jerripothula2016image}
Jerripothula, K.R., Cai, J., Yuan, J.: Image co-segmentation via saliency
  co-fusion. IEEE Transactions on Multimedia  \textbf{18}(9),  1896--1909
  (2016)

\bibitem{jing2019coarse}
Jing, L., Chen, Y., Tian, Y.: Coarse-to-fine semantic segmentation from
  image-level labels. IEEE Transactions on Image Processing  \textbf{29},
  225--236 (2019)

\bibitem{joulin2010discriminative}
Joulin, A., Bach, F., Ponce, J.: Discriminative clustering for image
  co-segmentation. In: 2010 IEEE Computer Society Conference on Computer Vision
  and Pattern Recognition. pp. 1943--1950. IEEE (2010)

\bibitem{kingma2014adam}
Kingma, D.P., Ba, J.: Adam: A method for stochastic optimization. arXiv
  preprint arXiv:1412.6980  (2014)

\bibitem{li2019group}
Li, B., Sun, Z., Li, Q., Wu, Y., Hu, A.: Group-wise deep object co-segmentation
  with co-attention recurrent neural network. In: Proceedings of the IEEE
  International Conference on Computer Vision. pp. 8519--8528 (2019)

\bibitem{li2020asif}
Li, C., Cong, R., Kwong, S., Hou, J., Fu, H., Zhu, G., Zhang, D., Huang, Q.:
  Asif-net: Attention steered interweave fusion network for rgb-d salient
  object detection. IEEE transactions on cybernetics  \textbf{51}(1),  88--100
  (2020)

\bibitem{li2018deep}
Li, W., Jafari, O.H., Rother, C.: Deep object co-segmentation. In: Asian
  Conference on Computer Vision. pp. 638--653. Springer (2018)

\bibitem{lin2016scribblesup}
Lin, D., Dai, J., Jia, J., He, K., Sun, J.: Scribblesup: Scribble-supervised
  convolutional networks for semantic segmentation. In: Proceedings of the IEEE
  Conference on Computer Vision and Pattern Recognition. pp. 3159--3167 (2016)

\bibitem{lin2014microsoft}
Lin, T.Y., Maire, M., Belongie, S., Hays, J., Perona, P., Ramanan, D.,
  Doll{\'a}r, P., Zitnick, C.L.: Microsoft coco: Common objects in context. In:
  European conference on computer vision. pp. 740--755. Springer (2014)

\bibitem{liu2020weakly}
Liu, W., Zhang, C., Lin, G., HUNG, T.Y., Miao, C.: Weakly supervised
  segmentation with maximum bipartite graph matching. In: Proceedings of the
  28th ACM International Conference on Multimedia. pp. 2085--2094 (2020)

\bibitem{liu2020crnet}
Liu, W., Zhang, C., Lin, G., Liu, F.: Crnet: Cross-reference networks for
  few-shot segmentation. In: Proceedings of the IEEE/CVF Conference on Computer
  Vision and Pattern Recognition. pp. 4165--4173 (2020)

\bibitem{long2015fully}
Long, J., Shelhamer, E., Darrell, T.: Fully convolutional networks for semantic
  segmentation. In: Proceedings of the IEEE conference on computer vision and
  pattern recognition. pp. 3431--3440 (2015)

\bibitem{mcintosh2018recurrent}
McIntosh, L., Maheswaranathan, N., Sussillo, D., Shlens, J.: Recurrent
  segmentation for variable computational budgets. In: Proceedings of the IEEE
  Conference on Computer Vision and Pattern Recognition Workshops. pp.
  1648--1657 (2018)

\bibitem{mukherjee2011scale}
Mukherjee, L., Singh, V., Peng, J.: Scale invariant cosegmentation for image
  groups. In: CVPR 2011. pp. 1881--1888. IEEE (2011)

\bibitem{mustafa2017semantically}
Mustafa, A., Hilton, A.: Semantically coherent co-segmentation and
  reconstruction of dynamic scenes. In: Proceedings of the IEEE Conference on
  Computer Vision and Pattern Recognition. pp. 422--431 (2017)

\bibitem{quan2016object}
Quan, R., Han, J., Zhang, D., Nie, F.: Object co-segmentation via graph
  optimized-flexible manifold ranking. In: Proceedings of the IEEE Conference
  on Computer Vision and Pattern Recognition. pp. 687--695 (2016)

\bibitem{ren2015faster}
Ren, S., He, K., Girshick, R., Sun, J.: Faster r-cnn: Towards real-time object
  detection with region proposal networks. In: Advances in neural information
  processing systems. pp. 91--99 (2015)

\bibitem{ronneberger2015u}
Ronneberger, O., Fischer, P., Brox, T.: U-net: Convolutional networks for
  biomedical image segmentation. In: International Conference on Medical image
  computing and computer-assisted intervention. pp. 234--241. Springer (2015)

\bibitem{rother2006cosegmentation}
Rother, C., Minka, T., Blake, A., Kolmogorov, V.: Cosegmentation of image pairs
  by histogram matching-incorporating a global constraint into mrfs. In: 2006
  IEEE Computer Society Conference on Computer Vision and Pattern Recognition
  (CVPR'06). vol.~1, pp. 993--1000. IEEE (2006)

\bibitem{rubinstein2013unsupervised}
Rubinstein, M., Joulin, A., Kopf, J., Liu, C.: Unsupervised joint object
  discovery and segmentation in internet images. In: Proceedings of the IEEE
  conference on computer vision and pattern recognition. pp. 1939--1946 (2013)

\bibitem{rubio2012unsupervised}
Rubio, J.C., Serrat, J., L{\'o}pez, A., Paragios, N.: Unsupervised
  co-segmentation through region matching. In: 2012 IEEE Conference on Computer
  Vision and Pattern Recognition. pp. 749--756. IEEE (2012)

\bibitem{sakinis2019interactive}
Sakinis, T., Milletari, F., Roth, H., Korfiatis, P., Kostandy, P., Philbrick,
  K., Akkus, Z., Xu, Z., Xu, D., Erickson, B.J.: Interactive segmentation of
  medical images through fully convolutional neural networks. arXiv preprint
  arXiv:1903.08205  (2019)

\bibitem{shi2015convolutional}
Shi, X., Chen, Z., Wang, H., Yeung, D.Y., Wong, W.K., Woo, W.c.: Convolutional
  lstm network: A machine learning approach for precipitation nowcasting. arXiv
  preprint arXiv:1506.04214  (2015)

\bibitem{shotton2006textonboost}
Shotton, J., Winn, J., Rother, C., Criminisi, A.: Textonboost: Joint
  appearance, shape and context modeling for multi-class object recognition and
  segmentation. In: European conference on computer vision. pp. 1--15. Springer
  (2006)

\bibitem{vicente2011object}
Vicente, S., Rother, C., Kolmogorov, V.: Object cosegmentation. In: CVPR 2011.
  pp. 2217--2224. IEEE (2011)

\bibitem{wang2013image}
Wang, F., Huang, Q., Guibas, L.J.: Image co-segmentation via consistent
  functional maps. In: Proceedings of the IEEE International Conference on
  Computer Vision. pp. 849--856 (2013)

\bibitem{wang2019zero}
Wang, W., Lu, X., Shen, J., Crandall, D.J., Shao, L.: Zero-shot video object
  segmentation via attentive graph neural networks. In: Proceedings of the IEEE
  International Conference on Computer Vision. pp. 9236--9245 (2019)

\bibitem{wang2018weakly}
Wang, X., You, S., Li, X., Ma, H.: Weakly-supervised semantic segmentation by
  iteratively mining common object features. In: Proceedings of the IEEE
  conference on computer vision and pattern recognition. pp. 1354--1362 (2018)

\bibitem{wei2017group}
Wei, L., Zhao, S., Bourahla, O.E.F., Li, X., Wu, F.: Group-wise deep
  co-saliency detection. arXiv preprint arXiv:1707.07381  (2017)

\bibitem{woo2018cbam}
Woo, S., Park, J., Lee, J.Y., So~Kweon, I.: Cbam: Convolutional block attention
  module. In: Proceedings of the European Conference on Computer Vision (ECCV).
  pp. 3--19 (2018)

\bibitem{yu2018learning}
Yu, C., Wang, J., Peng, C., Gao, C., Yu, G., Sang, N.: Learning a
  discriminative feature network for semantic segmentation. In: Proceedings of
  the IEEE conference on computer vision and pattern recognition. pp.
  1857--1866 (2018)

\bibitem{yu2018co}
Yu, H., Zheng, K., Fang, J., Guo, H., Feng, W., Wang, S.: Co-saliency detection
  within a single image. In: Proceedings of the AAAI Conference on Artificial
  Intelligence. vol.~32 (2018)

\bibitem{yu2019exemplar}
Yu, J.G., Li, Y., Gao, C., Gao, H., Xia, G.S., Yu, Z.L., Li, Y.: Exemplar-based
  recursive instance segmentation with application to plant image analysis.
  IEEE Transactions on Image Processing  \textbf{29},  389--404 (2019)

\bibitem{yuan2017deep}
Yuan, Z.H., Lu, T., Wu, Y.: Deep-dense conditional random fields for object
  co-segmentation. In: IJCAI. pp. 3371--3377 (2017)

\bibitem{zeiler2014visualizing}
Zeiler, M.D., Fergus, R.: Visualizing and understanding convolutional networks.
  In: European conference on computer vision. pp. 818--833. Springer (2014)

\bibitem{zhang2020deepemdv2}
Zhang, C., Cai, Y., Lin, G., Shen, C.: Deepemd: Differentiable earth mover's
  distance for few-shot learning (2020)

\bibitem{zhang2020deepemd}
Zhang, C., Cai, Y., Lin, G., Shen, C.: Deepemd: Few-shot image classification
  with differentiable earth mover's distance and structured classifiers. In:
  Proceedings of the IEEE/CVF Conference on Computer Vision and Pattern
  Recognition. pp. 12203--12213 (2020)

\bibitem{zhang2019pyramid}
Zhang, C., Lin, G., Liu, F., Guo, J., Wu, Q., Yao, R.: Pyramid graph networks
  with connection attentions for region-based one-shot semantic segmentation.
  In: Proceedings of the IEEE International Conference on Computer Vision. pp.
  9587--9595 (2019)

\bibitem{zhang2019canet}
Zhang, C., Lin, G., Liu, F., Yao, R., Shen, C.: Canet: Class-agnostic
  segmentation networks with iterative refinement and attentive few-shot
  learning. In: Proceedings of the IEEE Conference on Computer Vision and
  Pattern Recognition. pp. 5217--5226 (2019)

\bibitem{zhang2015cosaliency}
Zhang, D., Han, J., Han, J., Shao, L.: Cosaliency detection based on
  intrasaliency prior transfer and deep intersaliency mining. IEEE transactions
  on neural networks and learning systems  \textbf{27}(6),  1163--1176 (2015)

\bibitem{zhang2016detection}
Zhang, D., Han, J., Li, C., Wang, J., Li, X.: Detection of co-salient objects
  by looking deep and wide. International Journal of Computer Vision
  \textbf{120}(2),  215--232 (2016)

\bibitem{zhang2016co}
Zhang, D., Meng, D., Han, J.: Co-saliency detection via a self-paced
  multiple-instance learning framework. IEEE transactions on pattern analysis
  and machine intelligence  \textbf{39}(5),  865--878 (2016)

\bibitem{zhang2019mask}
Zhang, H., Tian, Y., Wang, K., Zhang, W., Wang, F.Y.: Mask ssd: An effective
  single-stage approach to object instance segmentation. IEEE Transactions on
  Image Processing  \textbf{29}(1),  2078--2093 (2019)

\bibitem{zhang2019deep}
Zhang, K., Chen, J., Liu, B., Liu, Q.: Deep object co-segmentation via
  spatial-semantic network modulation. arXiv preprint arXiv:1911.12950  (2019)

\bibitem{zhang2019rgb}
Zhang, Q., Huang, N., Yao, L., Zhang, D., Shan, C., Han, J.: Rgb-t salient
  object detection via fusing multi-level cnn features. IEEE Transactions on
  Image Processing  \textbf{29},  3321--3335 (2019)

\bibitem{zhang2020revisiting}
Zhang, Q., Xiao, T., Huang, N., Zhang, D., Han, J.: Revisiting feature fusion
  for rgb-t salient object detection. IEEE Transactions on Circuits and Systems
  for Video Technology  (2020)

\bibitem{zhao2017pyramid}
Zhao, H., Shi, J., Qi, X., Wang, X., Jia, J.: Pyramid scene parsing network.
  In: Proceedings of the IEEE conference on computer vision and pattern
  recognition. pp. 2881--2890 (2017)

\end{thebibliography}
}

\appendices




\ifCLASSOPTIONcaptionsoff
  \newpage
\fi

\end{document}


\section{More About Group Segmentation}

We present a more detailed analysis about group segmentation in this part. The group segmentation aims to predict the common objects in a group of images.
We apply different CNN based solutions in previous works to our network and then compare their performance.

Since the networks in most previous works for co-segmentation can only handle paired inputs, they often adopt various sampling strategies and fusion methods to undertake the group segmentation tasks, which can be applied to our network as well. An advantage of our network is that our network can handle more than two images at a time, as is described earlier, so the number of input images is flexible. However, directly sending all images in the group to our network is unrealistic due to the limited GPU memory. We therefore sample a small group of $k$ images at a time and send them to our network for predictions. Then, we fuse all the predictions of an image as the final mask. Suppose that there are $N$ images to be segmented in a group segmentation task and our network handles a tuple of $k$ images at a time. We compare the following strategies to undertake group segmentation:

\textbf{a)} Following \cite{banerjee2019cosegnet}, we sample all possible $k$-element tuples to make predictions and average all the corresponding confidence maps of an image as the final prediction. This method can make use of all images in the group to make predictions for each image.

\textbf{b)} Following \cite{zhang2019deep}, we randomly divide all images into $(N/k)$ small groups and each group has $k$ images. In this case, each image is only tested once and the whole evaluation process is quick.

\textbf{c)} Following \cite{li2018deep}, to predict the mask of an image, we randomly sample 5 groups of images from the other $N-1$ images, where each group has $k-1$ images. Then, the target image joins each of these groups to make a prediction with our network, and we average $5$ predictions to generate the final mask. The process is repeated for all  $N$ images.

\textbf{d)} Following \cite{wang2019zero}, to predict the mask of an image, we uniformly split the other ${N-1}$ images into ${T}$ groups, where ${T=(N-1)/(k-1) }$ and each group has $k-1$ images. Then, the target image joins each of these groups to make a prediction with our network, and we average $T$ predictions to generate the final mask. The process is repeated for all $N$ images.

Besides these solutions, there are also some previous works using graphs~\cite{quan2016object,han2017robust} or recurrent models~\cite{wang2019zero} to handle group segmentation, which, however, can not be applied to our network directly. Therefore, we compare the four solutions above with our network. We conduct experiments on the subset of Internet dataset, where each class has 100 images. 
For the solution \textbf{a}, generating all possible tuples for testing is unrealistic when $k$ is large. Therefore, we only test the cases of $k=2$ and $k=3$.
The result is shown in Table~\ref{tab:compare}. 
As we can see, predicting with more than two input images is consistently better than predicting with paired input images with all four strategies, which shows the superiority of our network design for group segmentation. The best tuple size is 4 or 5, while further increasing the tuple size no longer boosts the performance, as is observed.

\begin{table*}[]
\setlength{\abovecaptionskip}{5pt}
    \resizebox{\textwidth}{15mm}{
     \begin{tabular}{lccc|ccc|ccc|ccc}
 \Xhline{1pt}
\multicolumn{1}{c}{\multirow{2}{*}{\textbf{input-size} \ }} & \multicolumn{3}{c}{strategy a)}  & \multicolumn{3}{c}{strategy b)} & \multicolumn{3}{c}{strategy c)}& \multicolumn{3}{c}{strategy d)}  \\ 
\Xcline{2-13}{1pt}
\cline{2-13}
\multicolumn{1}{c}{}                        &  \multicolumn{1}{c}{Car\  } & \multicolumn{1}{c}{Airplane\  } & \multicolumn{1}{c}{Horse\ } & 
\multicolumn{1}{c}{Car\  } & \multicolumn{1}{c}{Airplane\  } & \multicolumn{1}{c}{Horse\ } & 
\multicolumn{1}{c}{Car\  } & \multicolumn{1}{c}{Airplane\  } & \multicolumn{1}{c}{Horse\ } & 
\multicolumn{1}{c}{Car\ } & \multicolumn{1}{c}{Airplane\ } & \multicolumn{1}{c}{Horse\ } &  \Xhline{1pt}
 ${k=2}$& 85.8 &77.2 &74.1 & 84.4& 78.3 & 73.9  &85.9 & 77.3  &74.0   & 85.7 & 77.3 &74.1  \\
 ${k=3}$ &86.6 &78.1 &75.0 & 84.8& 78.8 & 74.1  &86.6 & 78.1  &75.0  & 86.8 &78.5 &75.1 \\
 ${k=4}$&- &- &- &  85.6 & 78.9 & 75.1   &86.9 &78.6 &75.4  &\textbf{87.0} &78.8 & \textbf{75.5} \\
 ${k=5}$&- &- &- &  86.5 &  \textbf{79.3} & 75.4  &86.9 &78.6  &\textbf{75.5}  & \textbf{87.0} & 78.8 &\textbf{75.5}\\
 ${k=6}$&- &- &- & 86.4 & 79.0  &75.4 & 86.9 & 78.7 &75.4  & 86.9 & 78.7 & \textbf{75.5} \\
 ${k=7}$&- &- &- & 86.3& 78.6 &75.2  & 86.7 & 78.7  & 75.4 & 86.8 & 78.7 &   75.4   \\
 ${k=8}$&- &- &- &86.1& 78.9  & 75.3 & 86.7 & 78.7  & \textbf{75.5} & 86.8 & 78.7 &   \textbf{75.5}   \\
 \Xhline{1pt}
\end{tabular}}
    \caption{The results of group segmentation with different sampling strategies and different numbers of input images on the Internet dataset. Our network can handle more than two images at a time, which generate better results than our network with paired input images. }
    \label{tab:compare}
\end{table*}